%% file: main.tex
\documentclass[10pt,twocolumn,letterpaper]{article}

\usepackage{cvpr}              

\input{preamble}

\definecolor{cvprblue}{rgb}{0.21,0.49,0.74}
\usepackage[pagebackref,breaklinks,colorlinks,allcolors=cvprblue]{hyperref}

\usepackage{booktabs}

\usepackage{multirow}
\usepackage{adjustbox}
\usepackage{array}
\usepackage{wrapfig,lipsum}
\usepackage[font={footnotesize}]{caption}
\usepackage{makecell}

\usepackage{adjustbox}
\usepackage{graphicx}
\usepackage{fix-cm}
\makeatletter
\def\thickhline{%
  \noalign{\ifnum0=`}\fi\hrule \@height \thickarrayrulewidth \futurelet
   \reserved@a\@xthickhline}
\def\@xthickhline{\ifx\reserved@a\thickhline
               \vskip\doublerulesep
               \vskip-\thickarrayrulewidth
             \fi
      \ifnum0=`{\fi}}
\makeatother
\newlength{\thickarrayrulewidth}
\usepackage{makecell}

\usepackage{adjustbox}
\usepackage{graphicx}
\usepackage{fix-cm}
\makeatletter
\def\thickhline{%
  \noalign{\ifnum0=`}\fi\hrule \@height \thickarrayrulewidth \futurelet
   \reserved@a\@xthickhline}
   
\def\@xthickhline{\ifx\reserved@a\thickhline
               \vskip\doublerulesep
               \vskip-\thickarrayrulewidth
             \fi
      \ifnum0=`{\fi}}
\makeatother
\usepackage[dvipsnames]{xcolor}
\usepackage{pifont}
\usepackage{graphicx}
\usepackage{colortbl}
\usepackage{arydshln}
\usepackage{tikz}
\definecolor{Gray}{gray}{0.8}
\definecolor{LG}{gray}{.92}

\usepackage{setspace}
\usepackage{bm}
\usepackage{tikz}

\usepackage{enumitem}
\usepackage{lipsum} 

\usepackage{xcolor,colortbl}
\usepackage{amsmath}

\definecolor{Gray}{gray}{0.85}
\definecolor{LightCyan}{rgb}{0.88,1,1}
\newcolumntype{a}{>{\columncolor{Gray}}c}

\usepackage{multirow}
\usepackage{adjustbox}
\usepackage{array}
\usepackage{wrapfig,lipsum}

\usepackage{makecell}

\usepackage{adjustbox}
\usepackage{graphicx}
\usepackage{fix-cm}
\makeatletter
\def\thickhline{%
  \noalign{\ifnum0=`}\fi\hrule \@height \thickarrayrulewidth \futurelet
   \reserved@a\@xthickhline}
\def\@xthickhline{\ifx\reserved@a\thickhline
               \vskip\doublerulesep
               \vskip-\thickarrayrulewidth
             \fi
      \ifnum0=`{\fi}}
\makeatother

\usepackage{setspace}
\usepackage{bm}
\usepackage{tikz}

\usepackage{enumitem}

\definecolor{darkgreen}{rgb}{0.2, 0.7, 0.1}

\definecolor{Gray}{gray}{0.8}
\definecolor{LG}{gray}{.92}

\def\paperID{35696} 
\def\confName{CVPR}
\def\confYear{2026}

\title{Event6D: Event-based Novel Object 6D Pose Tracking} 

\author{
Jae-Young Kang$^{1}$\thanks{Equal contribution} \quad
Hoonhee Cho$^{1}$\footnotemark[1] \quad
Taeyeop Lee$^{1}$\footnotemark[1] \\
Minjun Kang$^{1}$ \quad
Bowen Wen$^{2}$ \quad
Youngho Kim$^{1}$ \quad
Kuk-Jin Yoon$^{1}$ \\
{\tt\small $^{1}$KAIST \quad $^{2}$NVIDIA}
}

\begin{document}
\maketitle
\input{sec_cr/0_abstract}    
\input{sec_cr/1_intro}
\input{sec_cr/2_related_works}
\input{sec_cr/3_methods}

\input{sec_cr/4_dataset}

\input{sec_cr/5_experiments}
\input{sec_cr/6_conclusion}

\clearpage
\setcounter{page}{1}
\maketitlesupplementary

In this supplemental document, we provide additional details about our datasets and the EventTrack6D method.
Specifically, we provide
\begin{itemize}
\item Details of the introduced EventBlender6D, EventHO3D, and Event6D datasets
in Sections~\ref{sec:blender6d}, \ref{sec:ho3d}, and \ref{sec:event6d}.

\item Details of the object assets and evaluation protocol in Section~\ref{sec:asset}.

\item Implementation details of the proposed method and other methods in Section~\ref{sec:detail}.

\item Experiments on additional datasets and methods, along with further analyses, qualitative results, and video demonstrations, in Section~\ref{sec:additional}.

\end{itemize}

\section{EventBlender6D Dataset}
\label{sec:blender6d}

EventBlender6D is a synthetic benchmark for 6D object pose estimation in dynamic scenarios, constructed at three difficulty levels (easy, medium, hard) according to the number of objects present in each scene. The easy setting contains single-object scenes with 1,033 sequences, whereas the medium setting includes 2–4 objects per scene and 2,066 sequences featuring collisions and mutual occlusions. The hard setting further increases the complexity to 5–10 objects per scene with 1,033 sequences. Each sequence comprises 120 frames recorded at 60 fps, resulting in 2-second clips that capture the full evolution of the scene, from initial object placement to free fall under gravity and eventual rest.

The dataset uses Google Scanned Objects (GSO)~\cite{downs2022google} with a balanced sampling strategy that ensures uniform representation across all models. Each object is assigned randomized material properties, including surface roughness and specular reflectance values between 0 and 1.0. Objects are initialized at random positions and orientations within the workspace, with collision checking to ensure valid starting configurations. The physics simulation uses realistic parameters with mass, friction coefficient, and damping values for stable dynamics.

Object motion is governed by realistic gravitational physics, where objects fall naturally, undergo collisions in multi-object scenes, and settle on the floor following physically-based dynamics. Camera motion follows a hemispherical orbital trajectory with azimuthal rotation completing 2.0 to 3.5 full revolutions per sequence, while elevation angles are constrained between 5° and 85°. The orbital radius is adaptively determined based on object bounding boxes, with scaling factors of 1.2–1.5 for easy mode and 1.5–2.0 for medium mode. Throughout the sequence, the camera continuously tracks a dynamically updated point of interest positioned at the median location of all objects, ensuring that the workspace remains centered in the field of view as objects descend under gravity.

To generate event data, we follow the protocol of video2events~\cite{Gehrig_2020_CVPR}. We first upsample the video frame rate using the method~\cite{reda2022film} described in their pipeline, and then synthesize events using ESIM~\cite{rebecq2018esim}. Following prior work~\cite{hamann2025etap, klenk2024deep}, we additionally adapt the generated events by applying random contrast sensitivities sampled from $\mathcal{U}(0.16, 0.34)$.

\begin{figure*}[p]
    \centering
    \includegraphics[width=0.99\linewidth] {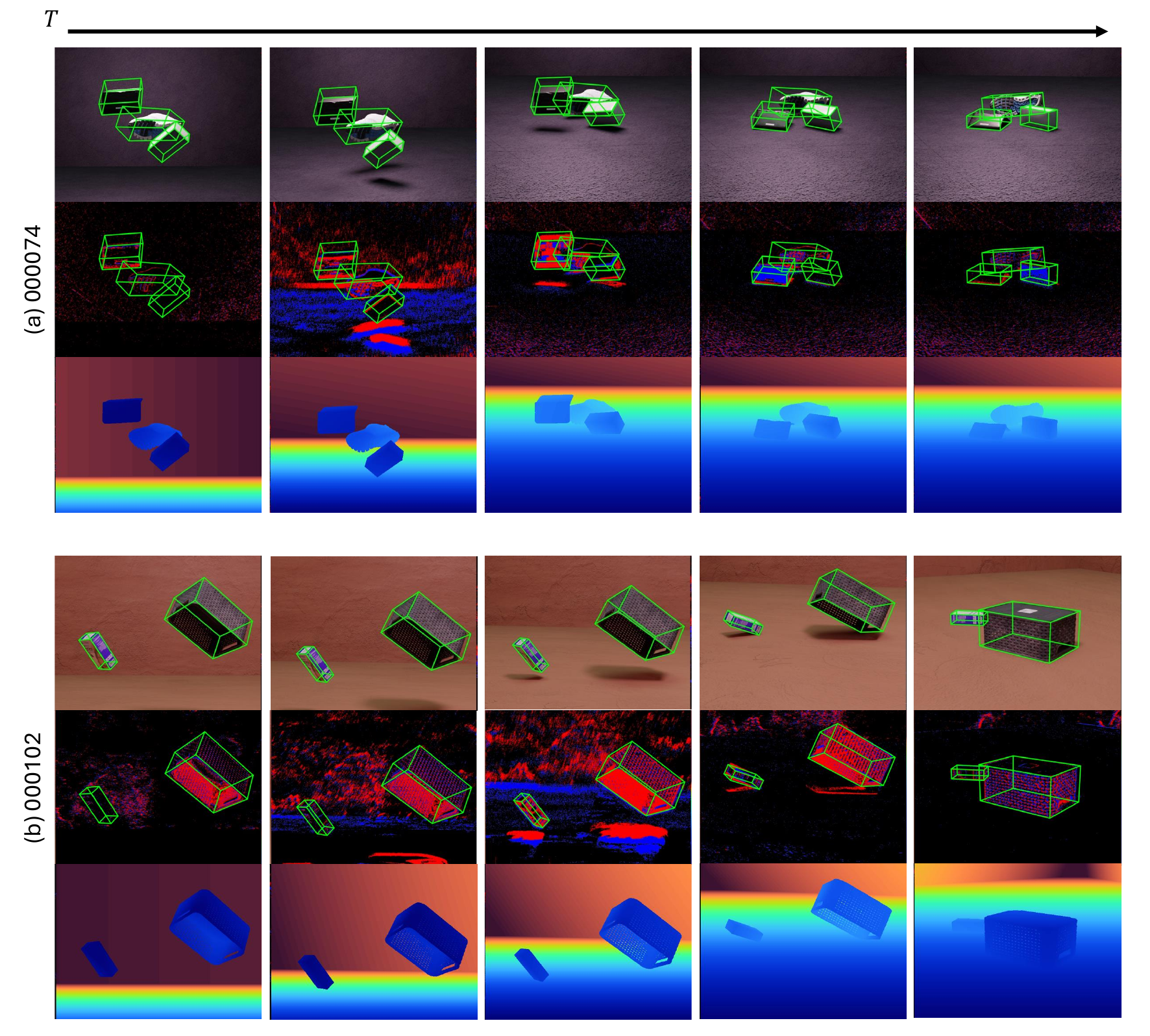}
    \vspace{-5pt}
    \caption{EventBlender6D samples visualized as temporal streams of RGB, event, depth, and corresponding 6D object poses.}
    \label{fig:dataset_blender}
\end{figure*}

Dataset samples are provided in Fig.~\ref{fig:dataset_blender}, and since the EventBlender6D data are rendered, the ground-truth 6D object poses are highly accurate.


\section{EventHO3D Dataset}
\label{sec:ho3d}
\begin{figure*}[p]
    \centering
    \includegraphics[width=0.99\linewidth]{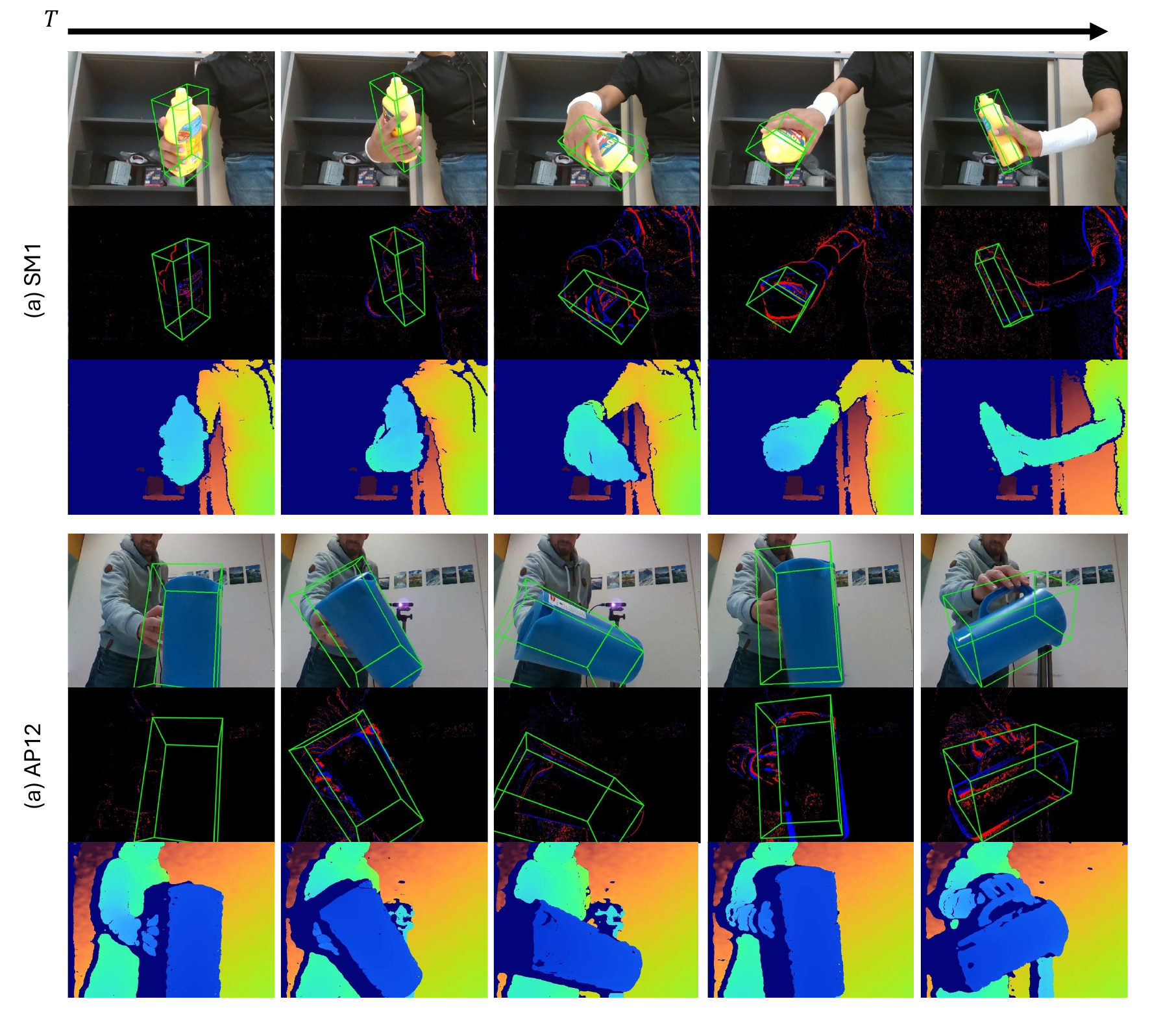}
    \vspace{-5pt}
    \caption{EventHO3D samples visualized as temporal streams of RGB, event, depth, and corresponding 6D object poses.
}
    \label{fig:dataset_ho3d}
\end{figure*}

For the HO3D dataset~\cite{hampali2020honnotate}, which consists of real-world markerless RGB-D hand–object interactions with 3D hand poses and 6D object poses obtained via sequence-level joint optimization, we generate event data using the same pipeline as EventBlender6D. Through this process, we construct the EventHO3D dataset. 
Note that EventHO3D is used only to assess the model’s generalization capability under diverse conditions, and none of its data are used for training.
Examples from the EventHO3D dataset are illustrated in Fig.~\ref{fig:dataset_ho3d}.


\section{Event6D Dataset}
\label{sec:event6d}
To acquire the Event6D dataset, we used three primary sensing systems: an RGB-D camera, an event camera, and an OptiTrack motion-capture system for providing ground-truth poses.
To reliably collect data from these heterogeneous sensors, two key procedures are required: cross-system calibration to align their coordinate frames, and time synchronization to ensure that all systems share a consistent temporal reference.
\subsection{Calibration}

\subsubsection{Camera Parameter Calibration}

\begin{figure}[t]
    \centering
    \includegraphics[width=1.0\linewidth]{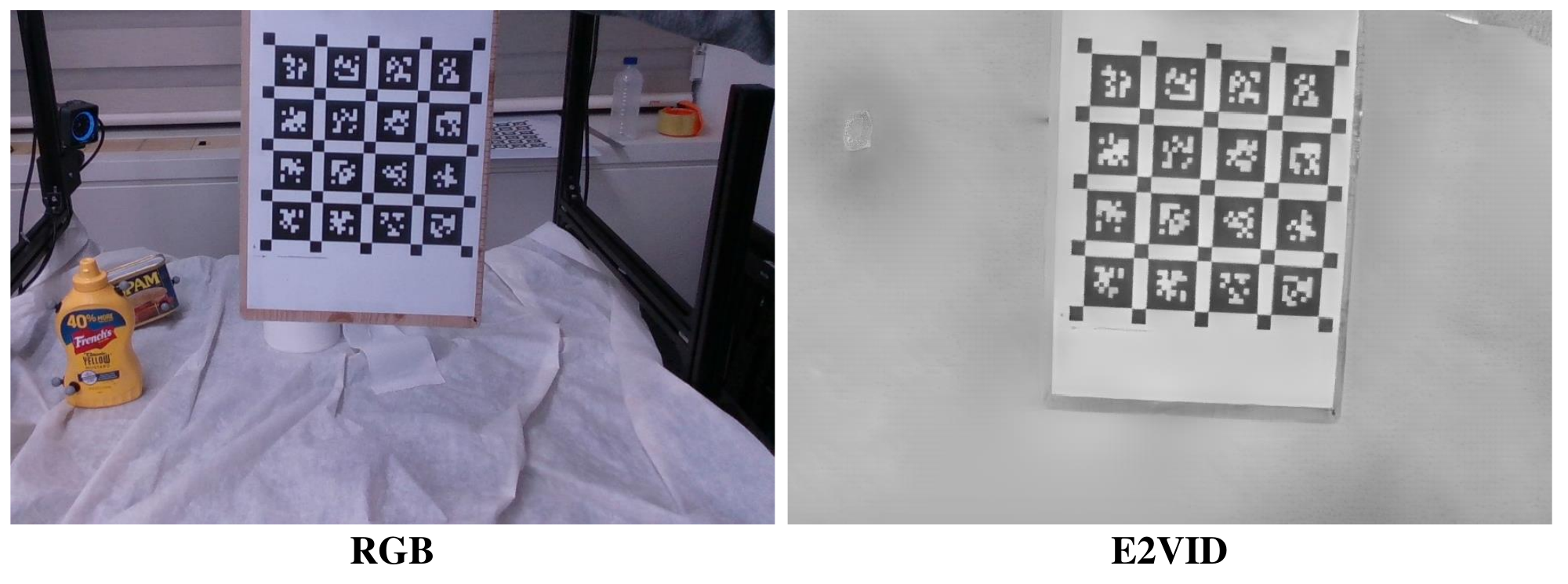}
    \vspace{-18pt}
    \caption{Examples of the data used for camera calibration.}
    \label{fig:intrinsic}
\end{figure}

Event cameras are inherently sparse and asynchronous, which makes their standalone calibration already challenging. Calibrating them jointly with conventional cameras is even more difficult. To address this, following prior works, we convert event streams into dense, temporally aligned images using a pretrained event-to-image reconstruction model.
As shown in Fig.~\ref{fig:intrinsic}, we reconstruct intensity images from the raw events using E2VID~\cite{e2vid}, and perform calibration on these reconstructed frames. For the calibration toolbox, we adopt Kalibr~\cite{furgale2013unified}, which is robust to noisy measurements and allows us to obtain both the intrinsic and extrinsic parameters of each camera.
Through this process, we obtain depth that is aligned with the event camera.

\begin{figure}[b]
    \centering
    \includegraphics[width=1.0\linewidth]{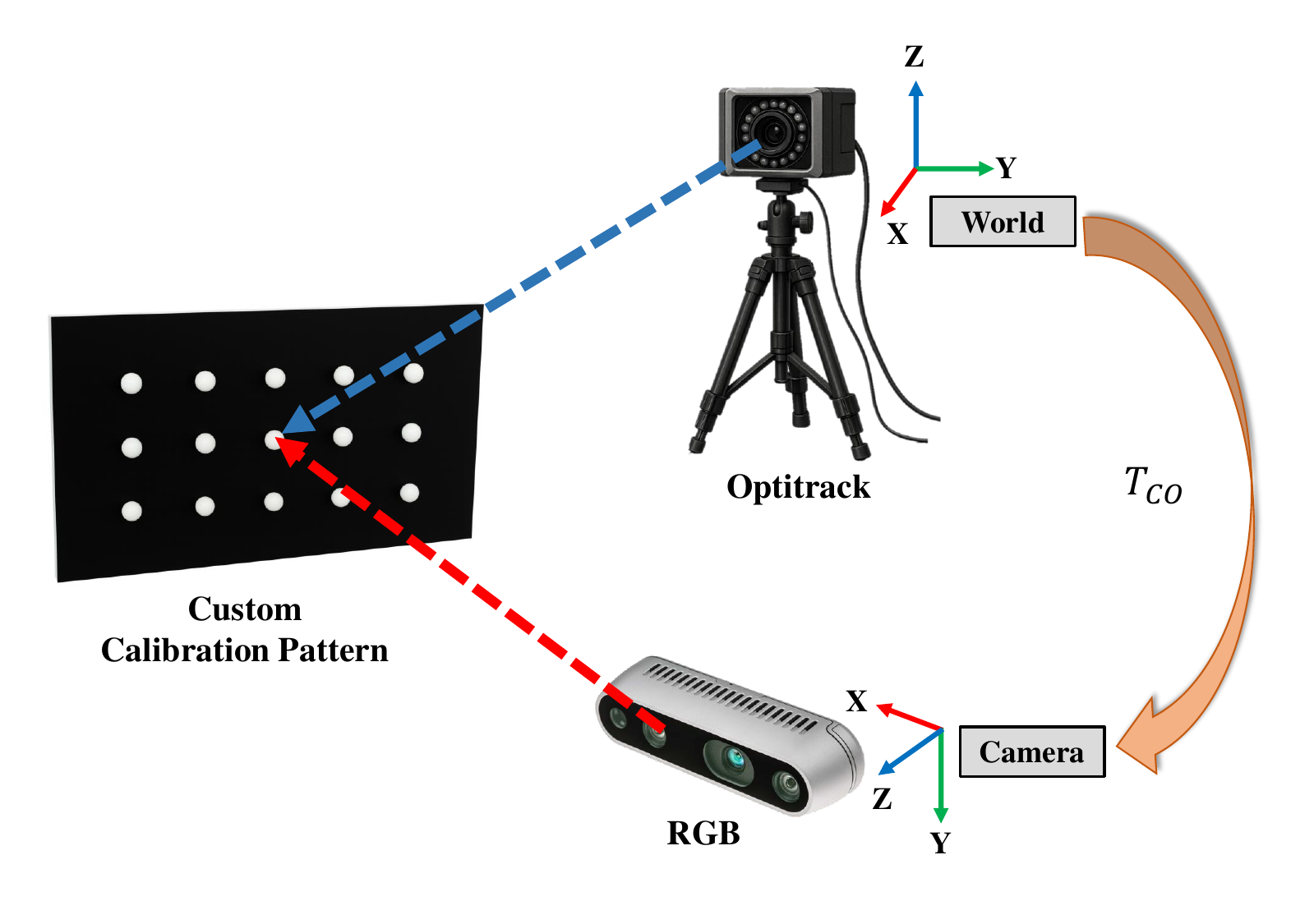}
    \caption{Illustration of Hand-eye calibration. We denote the OptiTrack (motion-capture) world coordinate frame as $O$ and the camera’s optical frame as $C$. The transformation from the OptiTrack frame to the camera frame is represented by $T_{CO}$.
}
    \label{fig:hadn_eye}
\end{figure}

\subsubsection{Hand-Eye Calibration}

Our objective is to estimate the 6D pose of each object in the camera coordinate frame. However, the OptiTrack motion-capture system provides measurements in its own world coordinate frame, which makes cross-system alignment essential. To bridge this gap, we estimate the transformation from the OptiTrack world frame to the camera coordinate frame by directly aligning the 2D observations in the camera images with the corresponding 3D points measured by the OptiTrack system. Specifically, we formulate the problem as a direct 2D--3D registration and solve it through a robust non-linear optimization procedure. This allows us to accurately map the OptiTrack world frame onto the camera coordinate frame and ensures that all subsequent 6D pose annotations are expressed consistently in the camera’s reference system.

\noindent
\textbf{Coordinate Frames.}
As shown in Fig.~\ref{fig:hadn_eye}, we denote the OptiTrack (motion-capture) world coordinate frame by $O$ and the camera’s optical frame by $C$. At each timestamp $t_m$, the OptiTrack system provides the 3D positions of the checkerboard corners $\mathbf{P}^O_n(t_m)$, where $n$ indexes individual checkerboard corners. The camera simultaneously observes the same corners in the image plane, yielding the corresponding 2D measurements $\mathbf{x}_{mn}$.

\noindent
\textbf{2D-3D Optimization.}
Our goal is to estimate the transformation from the OptiTrack world frame to the camera frame,
\begin{equation}
T_{CO} =
\begin{bmatrix}
R_{CO} & \mathbf{t}_{CO} \\
\mathbf{0}^\top & 1
\end{bmatrix},
\end{equation}
where $R_{CO}$ and $\mathbf{t}_{CO}$ denote rotation and translation, respectively.
Given camera intrinsics $K$, the predicted image projection of a 3D point is
\begin{equation}
\hat{\mathbf{x}}_{mn}
= \pi\left( K \, T_{CO} \, \mathbf{P}_n^O(t_m) \right),
\end{equation}
where $\pi(\cdot)$ denotes the perspective projection function.
We estimate $T_{CO}$ by minimizing the total reprojection error:
\begin{equation}
\label{eq:reprojection_opt}
\min_{R_{CO}, \mathbf{t}_{CO}}
\sum_{m,n}
\left\|
\mathbf{x}_{mn}
-
\hat{\mathbf{x}}_{mn}(R_{CO}, \mathbf{t}_{CO})
\right\|^{2}.
\end{equation}
This non-linear least-squares problem is solved via Levenberg-Marquardt algorithm.

\noindent
\textbf{RANSAC-based Outlier Rejection.}
To handle noisy 2D detections from the camera coordinate, we adopt a RANSAC scheme before the final refinement. At each iteration, a minimal subset of 2D-3D correspondences is sampled to compute a candidate pose $\hat{T}_{CO}$. The remaining correspondences are tested for inlier support:
\begin{equation}
\left\|
\mathbf{x}_{mn} - \hat{\mathbf{x}}_{mn}(\hat{T}_{CO})
\right\| < \tau,
\end{equation}
where $\tau$ is a reprojection error threshold.
The hypothesis with the largest inlier set is retained, and the final pose estimate is obtained by solving \eqref{eq:reprojection_opt} using only the inliers.

\noindent
\textbf{Ground-truth 6D Object Pose.}
The resulting optimized transformation $T_{CO}$ directly represents the camera pose with respect to the OptiTrack world coordinate frame. 
Since the object poses provided by OptiTrack are expressed in the OptiTrack world frame, we first transform them into the camera coordinate frame using $T_{CO}$. However, the resulting object pose centers are not perfectly aligned with the true centers of the corresponding CAD models. To address this, we obtain an initial 6D object pose by combining FoundationPose~\cite{wen2023foundationpose} with masks generated by the Segment Anything Model~\cite{kirillov2023segment}, and then manually refine this pose. We subsequently convert only this refined pose into the OptiTrack coordinate frame and use it as the ground-truth annotation.


\subsection{Trigger System}

\begin{figure}[h]
    \vspace{-5pt}
    \centering
    \includegraphics[width=.91\linewidth]{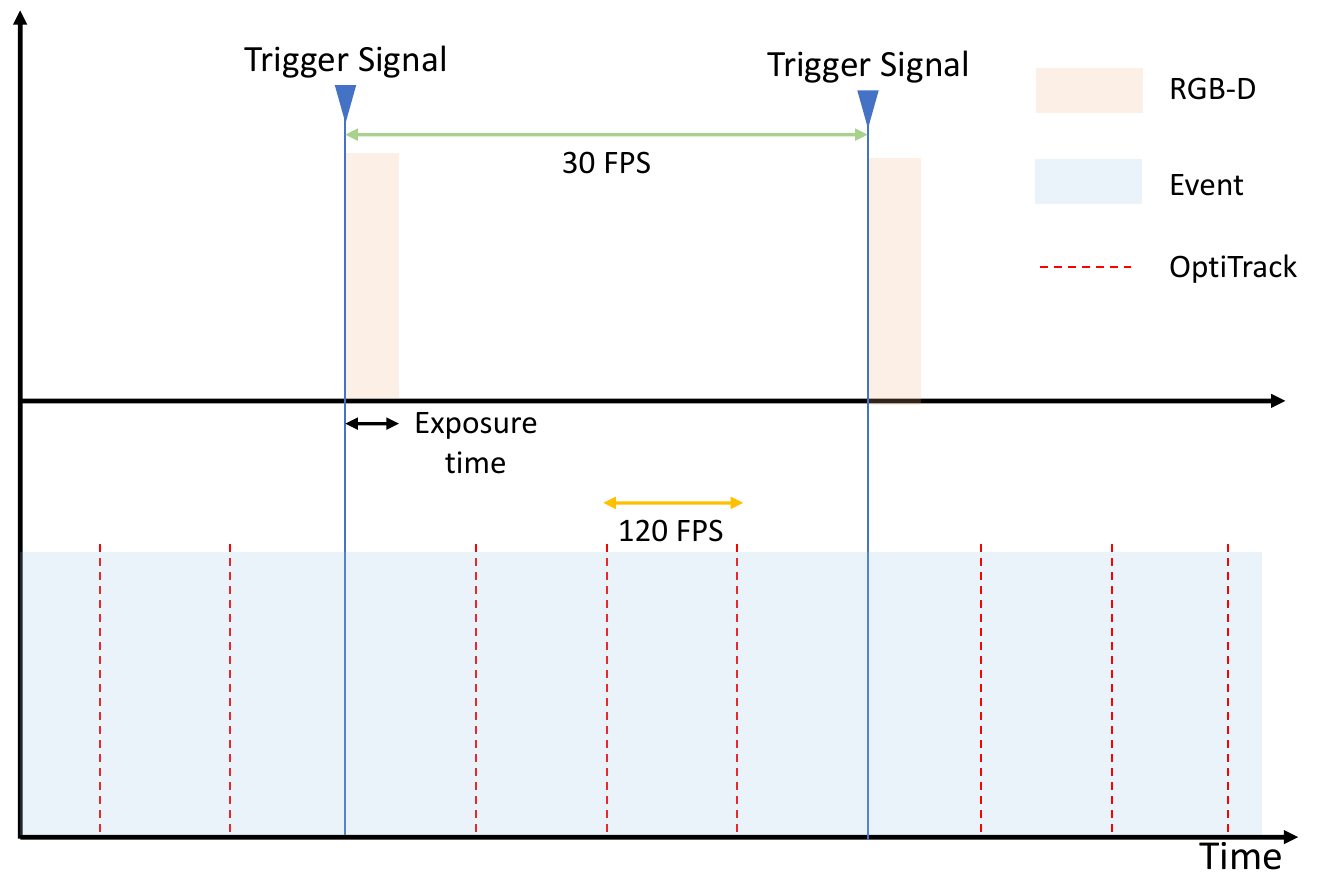}
    \vspace{-8pt}
    \caption{Visualization of trigger signals for overall system.
}
    \label{fig:trigger}
\end{figure}

To ensure that all data is captured in the same precise time domain, we employed a hardware trigger system to synchronize the acquisition times. The RGB-D camera used in our setup, the RealSense D435i, can internally generate external trigger signals at 30 FPS. These signals are then received and processed by both the event camera and the OptiTrack system. As illustrated in Fig.~\ref{fig:trigger}, the RGB-D camera captures data at 30 FPS and simultaneously outputs a trigger signal. Based on this trigger, the event camera can segment its event stream into slices, and the OptiTrack system can align its ground-truth acquisition with the same timing. Furthermore, OptiTrack can subdivide each external trigger interval into smaller segments using its internal multiplier, enabling ground-truth capture at 120 FPS, which is four times faster than the RGB-D camera rate.

\subsection{Dataset Details}

\begin{figure*}[p]
    \centering
    \includegraphics[width=0.99\linewidth] {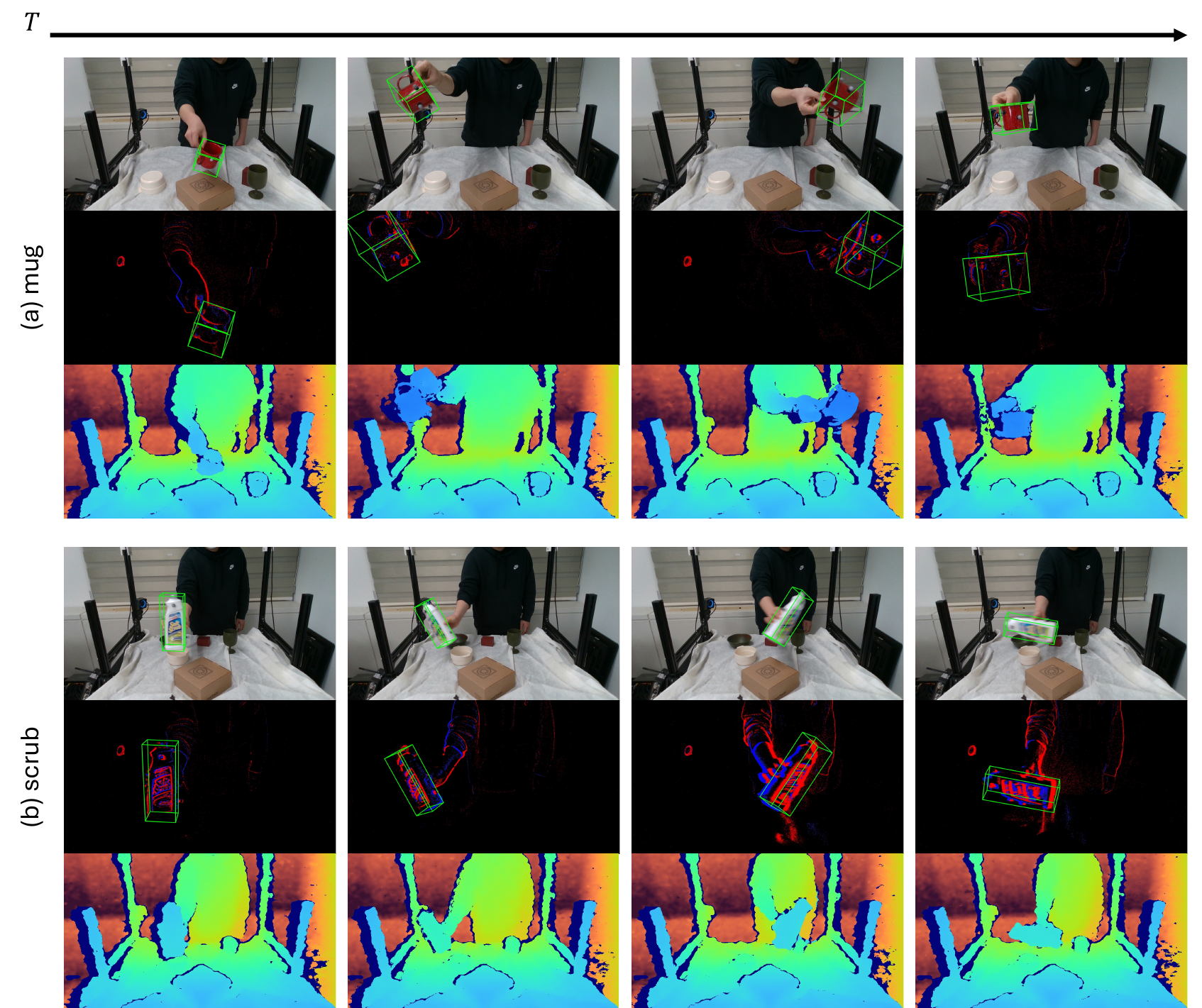}
    \vspace{-5pt}
    \caption{Event6D test samples visualized as temporal streams of RGB, event, depth, and corresponding 6D object poses.
}
    \label{fig:dataset_event6d}
\end{figure*}

We acquired the Event6D dataset such that each object exhibits dynamic, challenging, yet realistic motions. To this end, we designed the motions by imagining typical real-world usage of each object and mimicking the kinds of movements that would naturally occur. In our experiments, we only use the Event6D dataset as a test set and do not use the training split at all. However, Event6D differs from existing datasets in two major aspects: (i) it includes challenging and highly dynamic object motions, and (ii) it provides highly accurate ground-truth poses together with event and depth data. 
These aspects underscore the strengths of our Event6D dataset. Consequently, we also collected a training split to facilitate future research.
Detailed descriptions of the training and test sequences of the proposed Event6D dataset are provided in Table~\ref{tab:train_seq} and Table~\ref{tab:test_seq}, respectively, and representative dataset samples are illustrated in Fig.~\ref{fig:dataset_event6d}.

\section{Object Assets and Novel Object Evaluation}
\label{sec:asset}

\begin{figure}[t]
    \centering
    \includegraphics[width=1.0\linewidth]{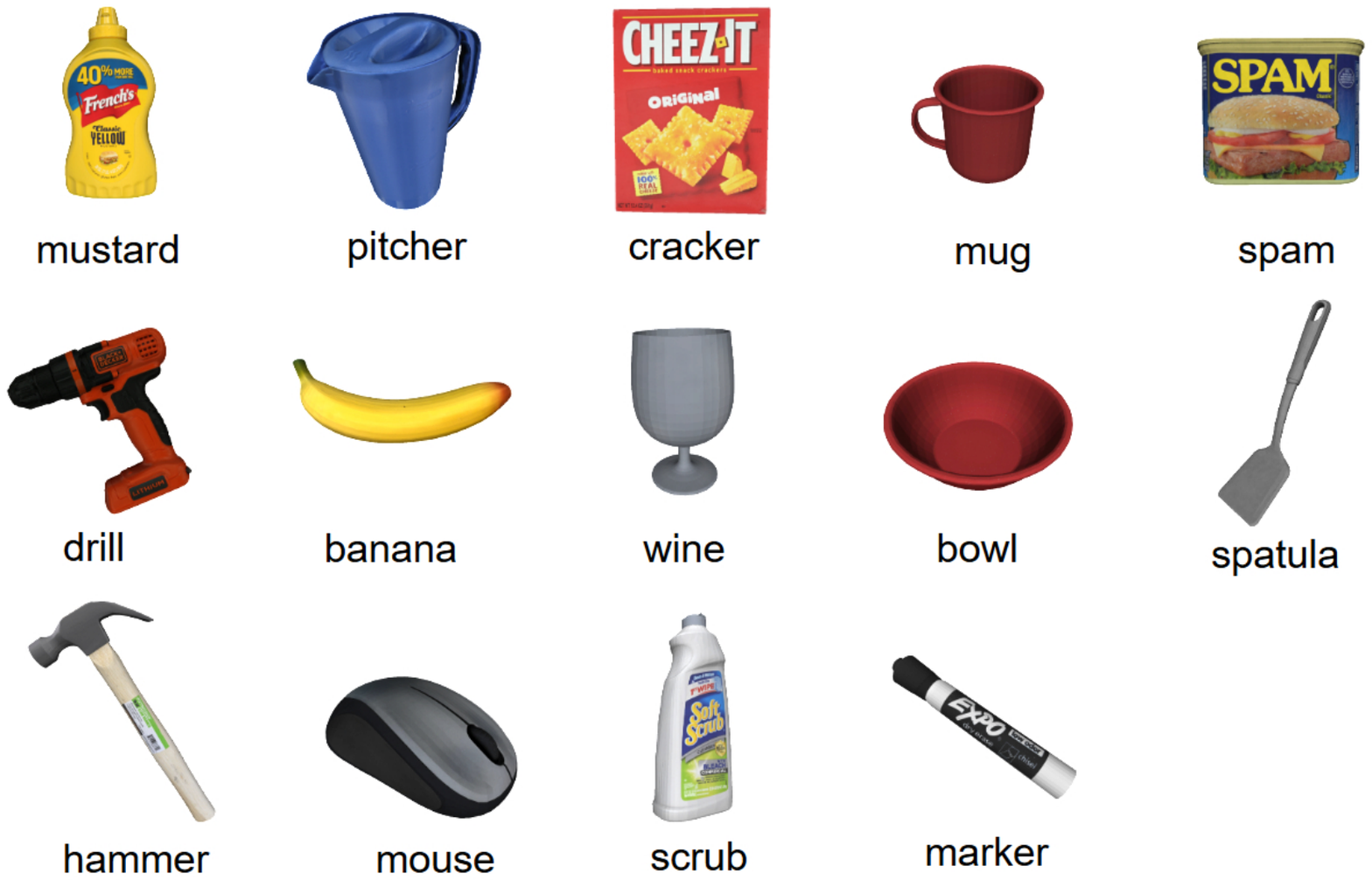}
    \vspace{-17pt}
    \caption{The object assets used in the Event6D dataset. The object assets do not overlap between EventBlender6D (used for training), ensuring proper novel-object testing.
}
    \label{fig:dataset}
\end{figure}

\textbf{Object Assets.}
For object assets, EventBlender6D consists of Google Scan Objects (GSO)~\cite{downs2022google}, which provides 1033 high-quality 3D scanned models with realistic textures.
The Event6D dataset consists of a subset of HOGrasp~\cite{cho2024dense} and Yale-CMU-Berkeley (YCB) dataset~\cite{calli2015ycb}, while HO3D consists of a subset of the YCB~\cite{calli2015ycb} dataset, as shown in Fig.~\ref{fig:dataset}.
For novel object pose estimation evaluation, we ensure that the training and test sets are strictly disjoint.
Specifically, the objects used in EventBlender6D (training) do not overlap with those in Event6D and EventHO3D (test), enabling rigorous evaluation of generalization to unseen objects.
In total, our dataset comprises 1047 unique object instances across diverse categories, including household items, tools, and objects relevant to manipulation.

\noindent \textbf{Object Instance Split for Train and Test.}
To evaluate novel object pose estimation capabilities, we maintain strict separation between training and evaluation objects.
The 1033 GSO objects in EventBlender6D serve as the training set, while Event6D and EventHO3D provide test scenarios with completely unseen objects from HOGrasp and YCB datasets.
This split ensures that models cannot rely on object-specific priors learned during training and must generalize to novel geometric and appearance characteristics.

\noindent  \textbf{CAD Model Acquisition.}
CAD models for GSO objects are directly obtained from the official repository with their provided high-quality meshes.
For YCB objects, we use the standardized CAD models from the official YCB Object and Model Set.
HOGrasp object meshes are either obtained from the original dataset or reconstructed using structure-from-motion techniques when high-quality CAD models are unavailable.
All meshes are preprocessed to ensure consistent coordinate frames, metric scale, and watertight geometry for physics simulation and rendering.
Fig.~\ref{fig:dataset} shows representative object assets from the Event6D dataset, illustrating the diversity of geometric complexity and visual appearance in our evaluation benchmark.

\section{Implementation Details}
\label{sec:detail}
\subsection{EventTrack6D}
For the event representation used in dual-modal reconstruction, we adopt a voxel grid~\cite{e2vid, zhu2019unsupervised, gehrig2021dsec} with a bin size of 5 for both the image and depth modalities. 
For training, we use two NVIDIA RTX A6000 GPUs and adopt a modular training strategy to improve stability. 
To effectively leverage prior knowledge learned from existing datasets, we initialize the image reconstruction module from a pretrained checkpoint~\cite{e2vid} and similarly initialize the refiner using a pretrained model~\cite{wen2023foundationpose}.
Specifically, we first train the dual-modal reconstruction as separate modules, with the image reconstruction module frozen, and then fine-tune the entire pipeline in an end-to-end manner, except for the LSTM parameters in the image reconstruction module, which are not further trained since they are not designed for sequential data.
We train our model using only the easy difficulty level of the EventBlender6D dataset, which already provide sufficient complexity and diversity for robust generalization across various scenarios.

\subsection{Event-based Baselines}

\subsubsection{Implementation Details of Each Model}

\noindent
\textbf{E2VID + MegaPose (MG).}
MegaPose (MG)~\cite{megapose} performs pose tracking on RGB or RGB-D images. We bridge the gap between event streams and image-based tracking by converting events to intensity images using E2VID~\cite{e2vid}. We use the pretrained MegaPose checkpoint, which has been trained on a diverse collection of datasets.

\noindent
\textbf{E2VID + FoundationPose (FP).}
FoundationPose (FP)~\cite{wen2023foundationpose} is a state-of-the-art RGB-D pose tracking method. Following the original implementation, we set the number of iterations in the FP pose refiner to 2.
To enable event-based tracking, we employ E2VID~\cite{e2vid} to reconstruct intensity images from event streams. These reconstructed images are fed into FP's RGB-D tracking pipeline, serving as our baseline configuration.
We utilize the FP checkpoint that has been pretrained on a large and diverse collection of datasets.

\noindent
\textbf{ETAP.}
We consider a hybrid formulation that combines the pre-trained event-based point tracking~\cite{hamann2025etap} with a rigid transformation–based update. Given the pose estimate from the previous timestamp, $\mathbf{T}_{t-\Delta t}$, we assume that the CAD model of the object is placed at the estimated pose, and then $n$ points are uniformly sampled from its 3D surface. These 3D points, $\mathbf{X}_{t-\Delta t}=\{\mathbf{X}^i = (x^i, y^i, z^i)\}_{i=1}^n$ are then projected onto the current event frame, resulting 2D pixel coordinates $\mathbf{x}_{t-\Delta t}=\{\mathbf{x}^i = K\mathbf{X}^i\}_{i=1}^n $, where $K$ represents a projection matrix of the event camera. 
We track the projected points using the event-based point tracker ETAP~\cite{hamann2025etap}, and denote the tracked 2D points as $\tilde{\mathbf{x}}_t$.
Depending on the availability of depth measurements, the final pose is computed using either a P$n$P~\cite{lepetit2009epnp} formulation or an ICP refinement~\cite{besl1992method_ICP}.

At time steps where depth measurements are available, the 3D coordinate of a tracked 2D point can be recovered from the depth map.  
Its depth is obtained by sampling the depth map $D_t$ at the tracked pixel location:
$d_t^i = D_t(u_t^i, v_t^i)$.
The 3D coordinate is then computed by back-projection:
\begin{equation}
    \mathbf{X}_t^i = d_t^i K^{-1} \tilde{\mathbf{x}}_t^i,
\end{equation}
where $\tilde{\mathbf{x}}_t^i = [u_t^i, v_t^i, 1]^\top$ is the tracked 2D point in homogeneous coordinates and $K$ denotes the camera intrinsic matrix. We align the previous 3D point cloud with the current observation using an ICP-based registration step.
\begin{equation}
    \Delta T_{t-\Delta t, t}=\mathrm{ICP}(\mathbf{X}_{t-\Delta t}, \mathbf{X}_t)
\end{equation}
\begin{equation}
    T_t= \Delta T_{t-\Delta t, t}T_{t-\Delta t}
\end{equation}

During time steps where no depth measurement is available, typically the interval between two consecutive depth inputs, we apply a P$n$P-based 2D–3D matching between the current 2D points and previously observed 3D points.
\begin{align}
    T_t=\mathrm{P}n\mathrm{P}(\mathbf{X}_{t-\Delta t}, \tilde{\mathbf{x}}_t)
\end{align}

\noindent
\textbf{Event-based FoundationPose.}
Since there are no existing learning-based event-driven methods that generalize well to novel objects, we train an adapted version of FoundationPose (FP)~\cite{wen2023foundationpose} that takes event data as input to serve as a strong event-based baseline. 
We build the training pipeline on top of the publicly available official implementation of FP, modifying the input interface to accept event voxel grids instead of RGB images. The network is initialized from the pretrained FP checkpoint, and training is carried out on the EventBlender6D dataset, using the same setup as for our method.

\subsubsection{Experimental Details at 120 FPS}

Unlike RGB-based models, the event-based baselines can perform inference at a higher temporal resolution, operating at 120 FPS as in our main experiments, rather than being limited to the 30 FPS of the depth stream. For MegaPose, FoundationPose, and our event-based adaptation of FP, we observe that they can still run without depth by masking the depth input with zeros. Based on this, we feed reconstructed images from E2VID in intervals where depth is not available and provide the depth input at timestamps where depth measurements are present. For the ETAP baseline, we use ICP-based tracking when depth is available, and fall back to a P$n$P-based pose update when only reconstructed image  information is present.

\section{Additional Experiments}
\label{sec:additional}

\subsection{Experiments on Other Datasets}
In addition to EventHO3D and Event6D, we further evaluate the trained model on the YCB-Ev dataset~\cite{rojtberg2025ycbev}, comparing our proposed EventTrack6D with the RGB+Depth-based FoundationPose (FP). 
Since the ground-truth (GT) annotations in YCB-Ev are generated using an RGB-D–based method, they can exhibit temporal inconsistencies within some sequences.
To mitigate this issue, we exclude such sequences from our evaluation. As shown in Table~\ref{tab:ycb_ev}, our method achieves higher quantitative scores than FP. However, we believe that these gains may not solely be attributed to the merits of our approach, but are also influenced by the fact that YCB-Ev does not employ hardware-level triggering, which can lead to misalignment between the different sensor streams. For this reason, we do not include the YCB-Ev results in the main paper and instead report them here for completeness.

We also considered conducting additional experiments on E-POSE~\cite{epose2025} and RGB-DE~\cite{dubeau2020rgb}, but were unable to do so because full public access to the necessary data is currently not available. In contrast, our Event6D dataset provides accurate ground-truth annotations using a motion capture system and ensures precise time synchronization across different modalities at the hardware level. This design highlights the reliability of Event6D as a benchmark, and we plan to maintain and release it as a well-curated public resource for the community.

\begin{table}[t]
\caption{Experiments on the YCB-Ev dataset.}
\vspace{-4pt}
\centering
\setlength{\tabcolsep}{13.5pt}
\resizebox{.47\textwidth}{!}{
\begin{tabular}{c||c|c}
\thickhline
Methods & FP~\cite{wen2023foundationpose} &  \textbf{EventTrack6D (Ours)} \\ 
\hline
Modality & RGB+Depth & Event+Depth\\
\thickhline
AR $\uparrow$ & 5.82 & 17.87 \\
\thickhline
\end{tabular}
}
\label{tab:ycb_ev}
\vspace{-4pt}
\end{table}

\subsection{Initialization Sensitivity}
Our method is designed to recover from rotation errors up to 20° and translation errors up to half the object diameter. Table~\ref{tab:initial_sensitivity} evaluates robustness to first-frame pose errors by applying $\Delta$ rotation and translation perturbations. Our method remains reliable within approximately 10$^\circ$ and 10 cm. 

\begin{table}[h]
\vspace{3pt}
\caption{Performance changes resulting from adding errors to the first-frame pose.
$\Delta 0^\circ$ and $\Delta 0$ cm indicate that no error was added.}
\vspace{-7pt}
\centering
\renewcommand{\arraystretch}{1.2}
\resizebox{.478\textwidth}{!}{
\setlength{\tabcolsep}{2pt}
\begin{tabular}{cc|cc|cc|cc|cc}
\thickhline
\multicolumn{2}{c|}{$\Delta 0^\circ$ \& $\Delta0$ cm} &  
\multicolumn{2}{c|}{$\Delta 3^\circ$ \& $\Delta 3$ cm} & \multicolumn{2}{c|}{$\Delta 5^\circ$ \& $\Delta 5$ cm} & \multicolumn{2}{c|}{$\Delta  10^\circ$ \& $\Delta 10$ cm}  & \multicolumn{2}{c}{$\Delta 15^\circ$ \& $\Delta15$ cm} \\
\hline
 ADD-S & ADD & ADD-S & ADD & ADD-S & ADD & ADD-S & ADD & ADD-S & ADD\\
\hline
52.79 & 25.26 &  50.74 & 23.75&	51.30&	23.69&	50.08&	23.38&	5.19&	1.81\\
\thickhline
\end{tabular}
}
\vspace{-12pt}
\label{tab:initial_sensitivity}
\end{table}

\subsection{Experiments with Other Existing Methods}
We first clarify that our task focuses on \textit{Novel Object 6D Pose Tracking}, where the model must track previously unseen objects during inference. Methods that rely on instance-level training do not fall within this scope. For example, RGB-D-E~\cite{dubeau2020rgb} methods are typically trained on a specific object instance and therefore do not exhibit the level of object generalization required for our setting.
LOPET~\cite{10645740} also presents challenges for our evaluation protocol. The method assumes line-based geometric priors and requires the target object to consist predominantly of linear structures. As illustrated in Fig.~\ref{fig:dataset}, many objects in our benchmark have curved or complex geometries, making it difficult to apply LOPET in a principled way. In addition, LOPET requires an initial line specification, which cannot be reliably provided for curved objects.
Although EDOPT~\cite{glover2024edopt} is not learning-based, it is capable of handling unseen objects and represents a valuable feature-based approach. However, our Event6D dataset contains objects moving at an average speed of 2,m/s, and, as noted in the official implementation, EDOPT is sensitive to rapid motion. In our experiments, the tracker quickly diverged under this dynamic setting, and we were therefore unable to obtain stable results suitable for reporting.

Given these considerations, we include the event-based FoundationPose~\cite{wen2023foundationpose} as a comparison method. FoundationPose has recently demonstrated strong generalization capabilities across novel objects and diverse scenarios. To ensure a fair and meaningful comparison within our event-based framework, we train an event-driven version of FoundationPose and use it as a competitive baseline in our evaluation.

\subsection{Qualitative Results}
\begin{figure*}[p]
    \centering
    \includegraphics[width=0.99\linewidth] {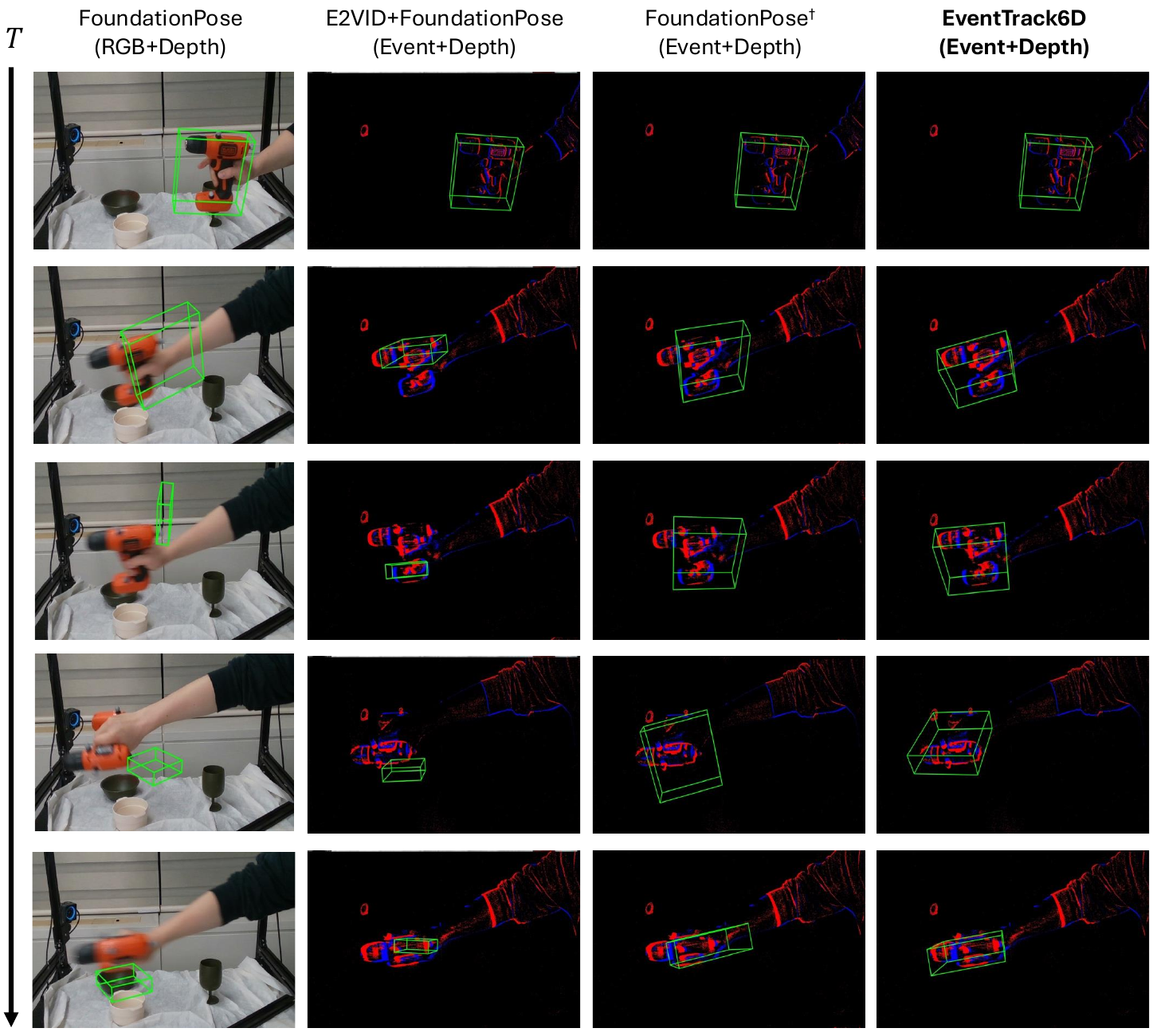}
    \vspace{-5pt}
    \caption{Qualitative comparison on the Event6D drill object sequence.  Although the event-based methods operate at intervals corresponding to 120 FPS, all visualizations are presented at the RGB frame rate of 30 FPS for consistency. $\dagger$ denotes that the model is trained with event inputs on the EventBlender6D dataset.
}
    \label{fig:qual_1}
\end{figure*}

\begin{figure*}[p]
    \centering
    \includegraphics[width=0.99\linewidth] {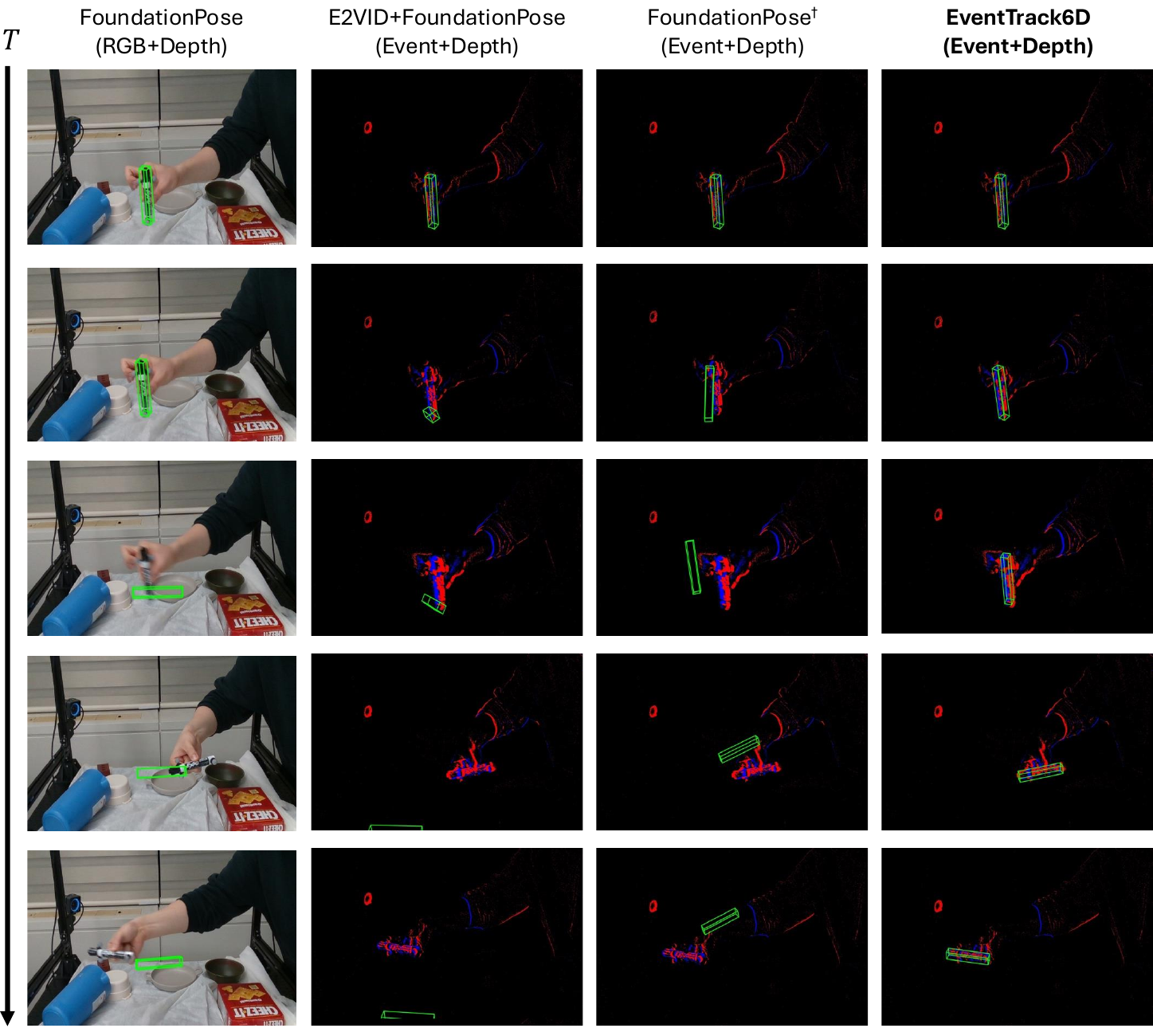}
    \vspace{-5pt}
    \caption{Qualitative comparison on the Event6D marker object sequence.  Although the event-based methods operate at intervals corresponding to 120 FPS, all visualizations are presented at the RGB frame rate of 30 FPS for consistency.$\dagger$ denotes that the model is trained with event inputs on the EventBlender6D dataset.}
    \label{fig:qual_2}
\end{figure*}

We provide additional qualitative results comparing the proposed EventTrack6D with several strong baselines: the state-of-the-art RGB-D tracker FoundationPose (FP)~\cite{wen2023foundationpose}, an E2VID~\cite{e2vid} + FP pipeline, and an event-adapted variant of FP that operates on event and depth inputs.
As can be seen in Figures~\ref{fig:qual_1} and \ref{fig:qual_2}, RGB-D-based FP quickly loses track of the object once the motion becomes large and highly dynamic. Moreover, using only E2VID for image reconstruction still provides insufficient geometric information, E2VID + FP often leading to additional tracking failures. The event-adapted FP variant is able to roughly follow the object motion, but it struggles to estimate the correct object scale and frequently produces inaccurate boundaries. In contrast, the proposed EventTrack6D accurately recovers the object pose even under highly dynamic motion, where RGB-D-based approaches face significant challenges. These results highlight that our method offers robust 6D tracking for novel objects in event-driven settings, providing a strong foundational baseline for future research on event-based object pose tracking.


\subsection{Video Demo in Dynamic Motion}
We additionally provide qualitative video results extracted from the test set to demonstrate that our method operates reliably over the time dimension. The accompanying demo includes highly dynamic and extreme motions in realistic scenes. Furthermore, we present additional videos recorded without the motion-capture system’s IR markers to showcase performance in even more realistic and challenging scenarios. As can be seen in these videos, the proposed method remains stable even under such extreme conditions, whereas RGB-D–based approaches often struggle to maintain accurate tracking.

\section{Acknowledgments}

This work was supported by the National Research Foundation of Korea(NRF) grant funded by the Korea government(MSIT) (RS-2026-25473963), the InnoCORE program of the Ministry of Science and ICT(26-InnoCORE-01), and the InnoCORE program of the Ministry of Science and ICT(N10250156).

\begin{table*}[p]
\caption{An overview of the proposed Event6D training set, which is released for future research and not used for training in our experiments. \textit{No. Frames} denotes the number of 30 FPS RGB and depth frames; the 6D poses are provided at 120 FPS.}
\vspace{-5pt}
\centering
\resizebox{1.0\textwidth}{!}{
\begin{tabular}{lcc}
\thickhline
Sequence Name & No. Frames & Description \\
\hline
\multicolumn{3}{l}{\textbf{Train Sequences}} \\
\hline
banana\_001 & 351 & The banana is moved by applying a translation at an appropriate speed. \\
banana\_002 & 294 & The banana is dynamically moved in all directions and orientations across 6-DoF. \\
banana\_003 & 320 & The banana is rapidly moved while varying its depth. \\
banana\_004 & 291 & The banana is moved rapidly by applying both rotation and translation. \\
bowl\_001 & 193 & The bowl is moved rapidly with translation and then rotated to include diverse motions. \\
bowl\_002 & 168 & The bowl is rapidly translated in multiple directions while its depth is quickly varied. \\
bowl\_003 & 166 & The bowl is rapidly rotated. \\
clamp\_001 & 244 & The clamp is rapidly rotated at various angles to include motion across all axes. \\
clamp\_002 & 328 & The clamp undergoes rapid combined rotational and translational motion as it is thrown and caught. \\
cracker\_001 & 266 & The cracker box is first translated rapidly and then rotated to enrich its motion. \\
cracker\_002 & 226 & The cracker box exhibits strong, rapid rotation, with translation occurring simultaneously. \\
cracker\_003 & 142 & The cracker box undergoes a throw-and-catch motion with rapidly and continuously varying depth. \\
cracker\_004 & 162 & The cracker box is repeatedly passed between both hands to generate dynamic motion. \\
drill\_001 & 493 & The drill undergoes rapid movement across diverse motion patterns. \\
drill\_002 & 408 & The drill is manipulated with abrupt, forceful movements, mimicking real drilling on various objects. \\
drill\_003 & 164 & The drill, placed among many objects, performs rotation-heavy motions that mimic drilling. \\
hammer\_001 & 119 & The hammer is repeatedly rotated by $180^\circ$ and driven through large translational motion. \\
marker\_001 & 201 & The marker is held in hand and moved rapidly with translational motion. \\
marker\_002 & 160 & The marker is held in hand and moved with larger, faster translational motion. \\
mouse\_001 & 257 & The mouse used for computers is held in hand and moved rapidly in various directions. \\
mug\_001 & 205 & The mug is moved rapidly with combined rotation and translation. \\
mug\_002 & 178 & The mug undergoes fast rotation and rapid motion, including collisions with another cup in a cheers gesture. \\
mustard\_001 & 215 & The mustard case undergoes rapid shaking and is translated over bowls. \\
pitcher\_001 & 289 & The pitcher is moved with rapid rotation, intermittently passed back and forth between both hands. \\
pitcher\_002 & 291 & The pitcher is used to rapidly pour water into multiple cups and bowls. \\
pitcher\_003 & 493 & The pitcher is used to rapidly pour water into multiple cups and bowls. \\
pudding\_001 & 151 & The pudding box undergoes extremely fast rotation while being moved. \\
pudding\_002 & 270 & The pudding box is spun at very high speed while being moved and placed on multiple bowls. \\
pudding\_003 & 364 & The pudding box is placed inside a bowl and shaken rapidly. \\
pudding\_004 & 271 & The pudding box moves with very fast translation and rotation, intermittently switching the holding hand. \\
scrub\_001 & 271 & The scrub cleanser bottle is rapidly translated and used to dispense cleanser onto multiple bottles. \\
scrub\_002 & 261 & The scrub cleanser bottle is held by its end and rotated widely with occasional hand switching. \\
spam\_001 & 661 & The spam can undergoes repeated rotations with varying depth, while the holding hand is switched. \\
spam\_002 & 221 & The spam can shows translation-dominant motion with repeated hand-to-hand throwing. \\
spatula\_001 & 224 & The spatula starts with fast motion and then performs cooking-like rotations perpendicular to the plane. \\
spatula\_002 & 261 & The spatula is driven quickly in a shaking motion, as if mixing something. \\
wine\_001 & 190 & The wine glass is moved dynamically with rotation-dominant motion at various angles, then placed on several bowls. \\
\hline
\multicolumn{1}{l}{Total} & 9,769 \\
\thickhline
\end{tabular}
}
\vspace{-10pt}
\label{tab:train_seq}
\end{table*}

\begin{table*}[p]
\caption{An overview of the proposed Event6D test set. \textit{No. Frames} denotes the number of 30 FPS RGB and depth frames; the 6D poses are provided at 120 FPS.}
\vspace{-6pt}
\centering
\resizebox{1.0\textwidth}{!}{
\begin{tabular}{lcc}
\thickhline
Sequence Name & No. Frames & Description \\
\hline
\multicolumn{3}{l}{\textbf{Test Sequences}} \\
\hline
banana & 301 & The banana is held by its stem and moved dynamically with both translation and rotation around that axis. \\
bowl & 261 & The bowl is moved with varying depth and rotated to reveal diverse viewpoints. \\
cracker & 308 & The cracker box is rotated through various angles and exchanged between both hands. \\
drill & 431 & The drill is moved quickly in a fixing-like action, performed at multiple orientations with repeated $180^\circ$ angle changes. \\
hammer & 246 & The hammer rapidly executes smashing motions as if breaking an object. \\
marker & 146 & The marker is rapidly moved with translation-dominant motion. \\
mouse & 192 & The mouse is held in hand and rapidly moved with rotation. \\
mug & 347 & The mug is grasped at the top and driven through wide and varied rotations.\\
mustard & 196 & The mustard bottle is tossed between both hands and moved back and forth over several bowls. \\
pitcher & 562 & The pitcher is rapidly rotated in one hand and then thrown and caught between both hands. \\
scrub & 276 & The scrub cleanser bottle undergoes multi-angle rotation while being moved with varying depth. \\
spam & 204 & The spam can undergoes dynamic 6-DoF movement involving both translation and rotation. \\
spatula & 263 & The spatula moves rapidly and includes stirring or flipping motions as in real cooking. \\
wine & 137 & The wine glass is rotated around the camera’s $z$-axis while undergoing depth variation. \\
\hline
\multicolumn{1}{l}{Total} & 3,870 \\
\thickhline
\end{tabular}
}
\vspace{-10pt}
\label{tab:test_seq}
\end{table*}

{
    \small
    \bibliographystyle{ieeenat_fullname}
    \bibliography{main}
}


\end{document}

%% file: preamble.tex
\definecolor{grey}{RGB}{230,230,230}
\definecolor{Ours}{RGB}{255,235,225}

\newcommand{\Tref}[1]{Table~\textcolor{blue}{\ref{#1}}}

\newcommand{\Fref}[1]{Fig.~\textcolor{blue}{\ref{#1}}}



%% file: sec_cr/0_abstract.tex
\begin{abstract}
Event cameras provide microsecond latency, making them suitable for 6D object pose tracking in fast, dynamic scenes where conventional RGB and depth pipelines suffer from motion blur and large pixel displacements. We introduce EventTrack6D, an event-depth tracking framework that generalizes to novel objects without object-specific training by reconstructing both intensity and depth at arbitrary timestamps between depth frames. Conditioned on the most recent depth measurement, our dual reconstruction recovers dense photometric and geometric cues from sparse event streams. Our EventTrack6D operates at over 120 FPS and maintains temporal consistency under rapid motion. To support training and evaluation, we introduce a comprehensive benchmark suite: a large-scale synthetic dataset for training and two complementary evaluation sets, including real and simulated event datasets. Trained exclusively on synthetic data, EventTrack6D generalizes effectively to real-world scenarios without fine-tuning, maintaining accurate tracking across diverse objects and motion patterns. Our method and datasets validate the effectiveness of event cameras for event-based 6D pose tracking of novel objects. Code and datasets are publicly available at \url{https://chohoonhee.github.io/Event6D}.

\end{abstract}

%% file: sec_cr/1_intro.tex
\section{Introduction}
\label{sec:intro}

Estimating 6D object pose is a fundamental problem in computer vision. Early 6D object pose estimation methods focused on instance-level approaches~\cite{xiang2017posecnn, zakharov-iccv19-dpod, wang2019densefusion}, where models are trained and evaluated on specific object instances. Research then progressed to category-level pose estimation~\cite{wang2019normalized, tian2020shape, lee2021category}, enabling generalization across object categories. 
Recent studies have advanced toward novel object generalization~\cite{nguyen2025bop, lin2024sam, gigaPose} in the context of robotic applications~\cite{lee2025delta}, developing models that handle unseen objects. However, accurate pose estimation from a single frame is insufficient for real-world applications requiring temporal consistency and continuous tracking. 
This motivates 6D object pose tracking~\cite{wen2023foundationpose, ponimatkin2025d,  wuthrich2013probabilistic}, which maintains consistent pose estimates across video sequences.

\begin{figure}[t]
    \centering
    \includegraphics[width=.478\textwidth,]{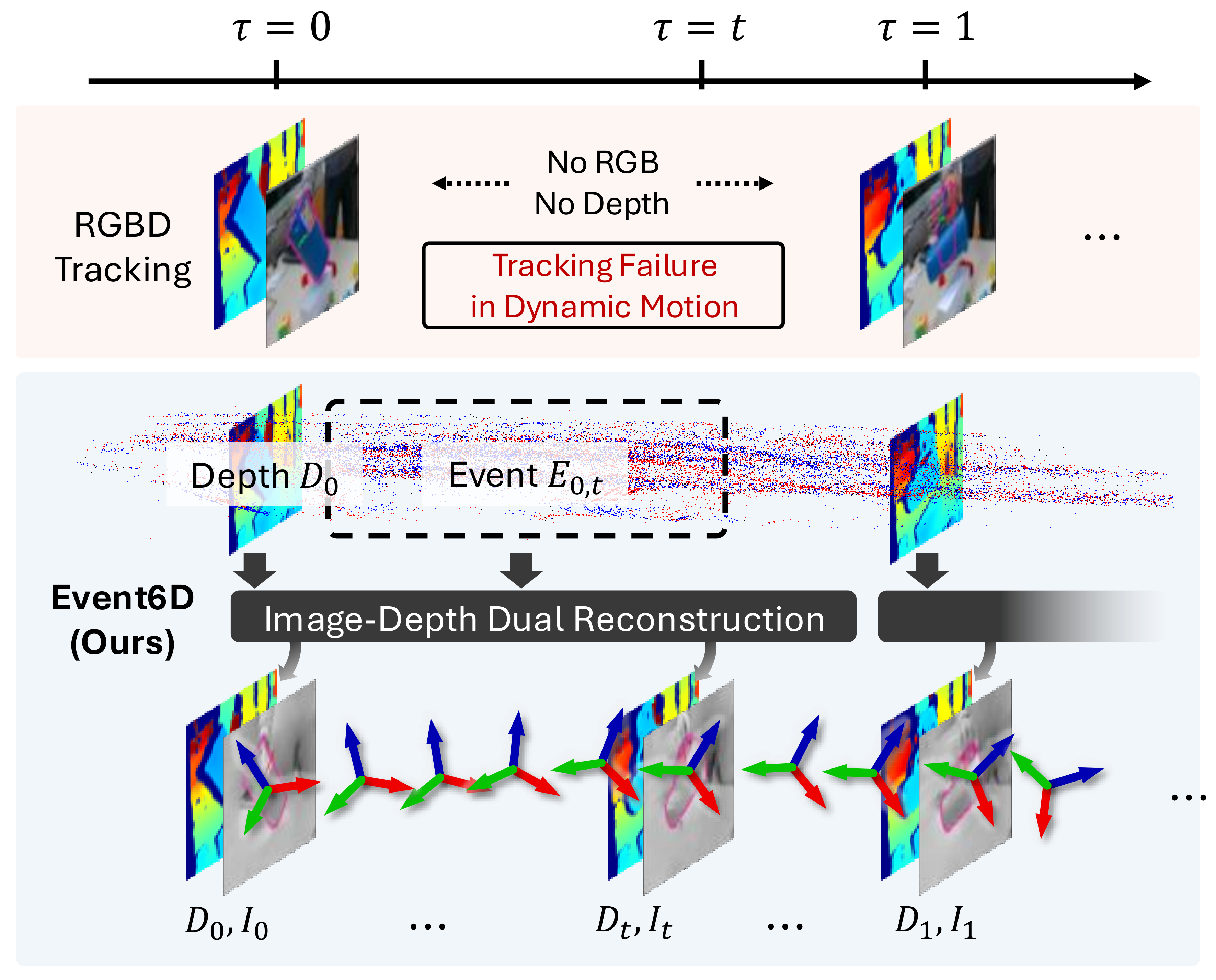}
    \vspace{-20pt}
    \caption{Conventional RGB-D based methods often fail under highly dynamic scenes due to limited frame rate from common RGB-D cameras. Our EventTrack6D addresses this issue by reconstructing dual modalities, image and depth, between consecutive depth frames to bridge the gap with event data. This enables inference at finer temporal intervals and yields robust tracking over highly dynamic motion.}
    \label{fig:teaser}
    \vspace{-10pt}
\end{figure}

\begin{table*}[t]
\centering
\caption{Dataset Comparison. Overview of publicly available event-based 6D object pose datasets. All datasets include RGB, Event, and depth data. n/a indicates that the paper does not provide the corresponding information.
}
\vspace{-3pt}
\resizebox{.95\textwidth}{!}{
\begin{tabular}{lcccccccc}
\thickhline
\multirow{2}{*}{Dataset}  & \multirow{2}{*}{Events} & \multirow{2}{*}{\# Samples} & \multirow{2}{*}{\# Objects} & Event & Annotation  & \multirow{2}{*}{Motion} & 6D Pose \\ 
& & & & Resolution &  Frequency (Hz) & & Annotation\\ 
\hline
YCB-Ev~\cite{rojtberg2025ycbev} & Real & 13,851 & 21 & 1280$\times$720 & 30  & Static & CosyPose~\cite{labbe2020cosypose}+ICG~\cite{stoiber2022iterative} \\
E-POSE~\cite{epose2025} &  Real & 333,357 & 13 & 346$\times$260 & 100 & Moderate & Registration~\cite{park2017colored}+ICP~\cite{besl1992method_ICP} \\
RGB-DE~\cite{dubeau2020rgb} & Real & 2,500 & 1 & 346$\times$260 & 30 & Dynamic & Manual+ICP~\cite{besl1992method_ICP}\\
\hline
\cellcolor{grey}\textbf{EventBlender6D~(Ours)}  & \cellcolor{grey}Synthetic & \cellcolor{grey}495,840  & \cellcolor{grey}1033  & \cellcolor{grey}640$\times$480   & \cellcolor{grey}60  & \cellcolor{grey}Dynamic & \cellcolor{grey}BlenderProc~\cite{blnederproc} \\
\cellcolor{grey}\textbf{EventHO3D~(Ours)} & \cellcolor{grey}Synthetic & \cellcolor{grey}103,462 & \cellcolor{grey}5 & \cellcolor{grey}640$\times$480  & \cellcolor{grey}n/a & \cellcolor{grey}Moderate & \cellcolor{grey}Multi-view Opt.~\cite{hampali2020honnotate} \\
\cellcolor{grey}\textbf{Event6D~(Ours)} & \cellcolor{grey}Real & \cellcolor{grey}54,556 & \cellcolor{grey}14 & \cellcolor{grey}1280$\times$720 & \cellcolor{grey}120  & \cellcolor{grey}Dynamic & \cellcolor{grey}Motion Capture 
~\cite{furtado2019comparative}\\
\thickhline
\label{tab:dataset_comparison}
\vspace{-16pt}
\end{tabular}}
\end{table*}

Despite recent advances in novel object 6D pose tracking, existing methods~\cite{wen2023foundationpose, moon2025co, nguyen2025gotrack, megapose} and datasets~\cite{hodan2024bop,wang2024ho,guo2023handal, calli2015ycb, chao2021dexycb, banerjee2024introducing} rely on RGB or depth modalities at conventional frame rates (up to 30 FPS), limiting their applicability to dynamic scenes with fast motions where motion blur and large pixel displacements degrade performance. This limitation motivates the need for robust 6D pose tracking methods in high-speed scenarios.

Event cameras~\cite{gallego2020event} emerged as a promising sensor for high-speed visual perception. Unlike conventional frame-based cameras~\cite{keselman2017intel, haggag2013measuring} that capture full images at fixed intervals, event cameras asynchronously record per-pixel brightness changes with microsecond temporal resolution, offering negligible motion blur, low latency, and high dynamic range.
Several works~\cite{rojtberg2025ycbev, epose2025, dubeau2020rgb} have investigated event-based 6D pose tracking tasks. RGB-DE~\cite{dubeau2020rgb} pioneered this area by introducing a single-object tracking dataset with an RGB-Depth-Event fusion method for instance-level 6D object pose tracking. 
However, their 6D pose annotations are limited to 30 Hz due to the RGB-D conventional frame rate. 
They show promising results by introducing RGB-Depth-Event fusion method for instance-level 6D object pose tracking.
As summarized in Table~\ref{tab:dataset_comparison}, recent datasets, such as YCB-Ev~\cite{rojtberg2025ycbev} and E-Pose~\cite{epose2025}, introduce multi-object event-based 6D datasets but still remain constrained to controlled settings with moderate or static object motion. Moreover, their annotation pipelines are limited in handling dynamic objects, as they rely on existing RGB-D 6D pose methods (CosyPose~\cite{labbe2020cosypose} and ICG~\cite{stoiber2022iterative}, or point cloud registration~\cite{park2017colored} with refinements~\cite{besl1992method_ICP}), which struggle with highly dynamic motions. 
Existing 6D event datasets are still limited in both scale and motion diversity, which are crucial for developing generalizable tracking methods.

To address these limitations, we propose EventTrack6D, an event-based 6D pose tracking framework that achieves robust tracking in high-speed scenarios and generalizes to novel unseen object instances without retraining. 
As shown in Fig.~\ref{fig:teaser}, our key idea is to reconstruct both intensity and depth at arbitrary timestamps between depth frames by leveraging event data conditioned on the most recent depth measurement. The reconstructed intensity and depth enable matching against CAD renderings, recovering photometric and geometric cues from sparse event data.
This dual reconstruction provides dense geometric and photometric information for render-and-compare objectives~\cite{wen2023foundationpose, megapose}, allowing pose estimation at temporal resolutions beyond the native depth frame rate. 
EventTrack6D runs at over 120 FPS with a lightweight architecture and maintains robust tracking in highly dynamic scenes.

Alongside our method, we introduce a comprehensive dataset suite: EventBlender6D for large-scale synthetic training (495k samples, 1k objects), Event6D (real-world, motion-captured) and EventHO3D (simulated event) for evaluation. By training exclusively on synthetic data, we demonstrate strong cross-domain generalization to real-world scenarios without fine-tuning.

The core contributions can be summarized as follows:
\begin{itemize}[noitemsep, topsep=0pt]
\item We propose EventTrack6D, an event camera-based 6D object pose tracking framework that generalizes to novel objects without retraining. 

\item We introduce a dual reconstruction approach that leverages event streams to reconstruct both intensity and depth between consecutive depth frames, which can be seamlessly integrated with the downstream module in a render-and-compare paradigm.

\item We present large-scale synthetic and real-world event-based datasets (EventBlender6D, EventHO3D, and Event6D) for both training and evaluating event camera-based 6D object pose tracking (see Table~\ref{tab:dataset_comparison}).
\end{itemize}

%% file: sec_cr/2_related_works.tex
\section{Related Works}
\label{sec:related_work}

\noindent
\textbf{6D Object Pose Estimation and Refinement.} 6D object pose estimation has been divided into instance-level~\cite{xiang2017posecnn, peng2019pvnet, wang2019densefusion}, category-level~\cite{wang2019normalized, lee2021category, lee2023tta}, and novel object pose estimation~\cite{wen2023foundationpose, gigaPose, lee2025any6d}. To improve initial pose predictions, pose refinement methods~\cite{zakharov2019dpod, megapose, genflow, li2018deepim, moon2025co} have been developed for each setting. 
However, most existing methods use RGB or depth as inputs and are limited by the sensing speed of common cameras. Under large and fast motion, these cameras produce severe motion blur in RGB frames and large inter-frame displacements, significantly degrading the performance.

\noindent

\textbf{6D Object Pose Tracking.}
Classical 6DoF tracking, including keypoint based~\cite{Rublee2011,Ozuysal2006,Rosten2005,Skrypnyk2004,Vacchetti2004}, edge based~\cite{Seo2014,Comport2006,Drummond2002b,Harris1990,stoiber2022srt3d}, and direct optimization methods~\cite{Seo2016,Crivellaro2014,Benhimane2004,Lucas1981,tian2022large}, struggles with textureless objects, clutter, and generalization. This has motivated learning based approaches~\cite{majcher20203d,deng2021poserbpf,wang2023deep,garon2017deep,wen2020se}, which often require large object specific datasets~\cite{wang2023deep,back2025graspclutter6d}. Category-level~\cite{wang20206,lin2022keypoint}, model-based~\cite{stoiber2022iterative,wuthrich2013probabilistic,issac2016depth}, and model-free trackers~\cite{wen2021bundletrack,wen2023bundlesdf} still depend on instance-level supervision, while recent work explores unseen object tracking~\cite{nguyen_pizza_2022} and iterative refinement~\cite{li2018deepim,megapose,genflow}. In this context~\cite{megapose,nguyen2025gotrack,wen2023foundationpose}, our EventTrack6D targets tracking of novel objects given CAD models and leverages event cameras for robustness under rapid and large motions.

\begin{figure*}[t]
    \centering\includegraphics[width=.99\textwidth,]{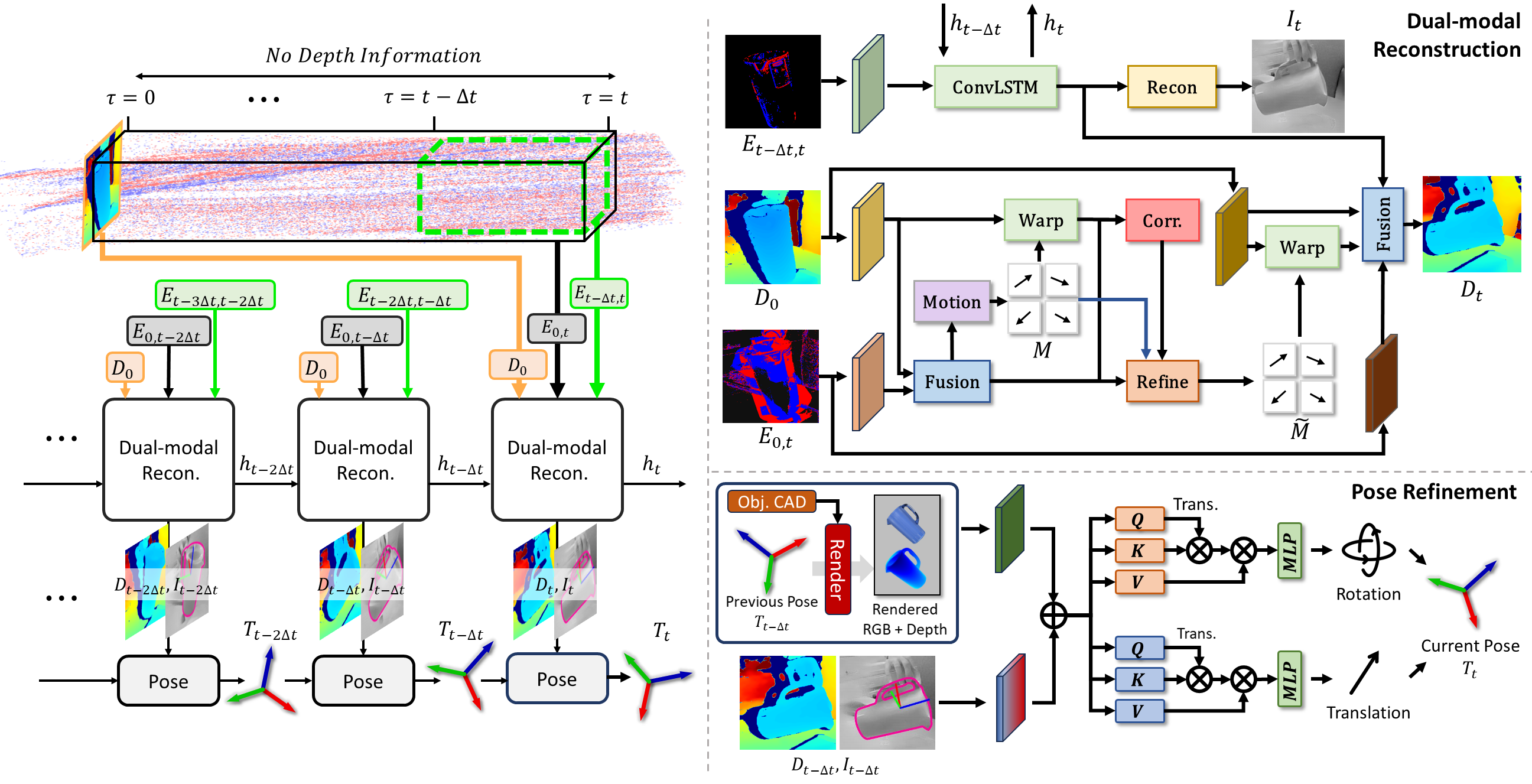}
    \vspace{-7pt}
    \caption{Overview of our EventTrack6D. EventTrack6D consists of a dual-modal reconstruction module and a pose refinement module. It can perform 6D pose tracking over high-frequency event stream, despite the limited frame rate of depth images which results in missing depth information between time intervals $\tau=0$ and $\tau=t$. 
    To achieve this, the dual-modal reconstruction module takes as input the most recent depth frame $D_0$, the event stream $E_{0,t}$ accumulated from that frame to the current time $t$ where depth frame is missing, as well as the event stream \(E_{t-\Delta t,t}\) from the most recent dual-modal reconstruction to the current time. From these inputs, it reconstructs the current intensity image $I_t$, and depth $D_t$. These reconstructed modalities are then used in a pose refinement module to estimate the 6D pose transformation from time $t\!-\!\Delta t$ to $t$.
}
    \label{fig:overall_framework}
    \vspace{-10pt}
\end{figure*}

\noindent
\textbf{Event-based Image Reconstruction.}
Event-based image reconstruction is a well-established topic in previous works~\cite{kim2008simultaneous, kim2016real, bardow2016simultaneous, reinbacher2016real, scheerlinck2019continuous, cook2011interacting,gehrig2018asynchronous,wang2021dual}. Recent work uses deep models to produce high-quality reconstructions~\cite{rebecq2019high, e2vid, scheerlinck2020fast, weng2021event, xiao2025event}, though supervised approaches typically require precisely aligned image–event pairs. Alternative paradigms have also been explored~\cite{wang2019event, pini2018learn, paredes2021back, fox2024unsupervised}, each introducing its own constraints. 
Notably, reconstructed images have been used for downstream tasks by leveraging dense photometric cues that sparse events lack~\cite{wang2021dual,wang2021joint}.

\noindent
\textbf{Event-based Depth Reconstruction.}
Depth estimation using event cameras has progressed rapidly~\cite{hidalgo2020learning,hu2025ede,liu2025high,zhu2025depth, kang2025temporal, cho2024temporal}, yet many methods are too slow for real-time or struggle with absolute scale. Rather than regressing depth directly from events, we exploit motion cues in the event stream~\cite{wan2025event,shen2025blinktrack,shiba2022secrets,gehrig2021raft,luo2024efficient} and the high frame rate of event sensing~\cite{kang2025unleashing,cho2025ev,gehrig2024low}. Inspired by design principles of event-based video frame interpolation methods~\cite{zhu2024video,tulyakov2021time,tulyakov2022time,zhang2022unifying,sun2023event,ma2024timelens,chen2025repurposing, cho2024tta}, we propose an event-driven depth extrapolation that predicts the current depth from incoming events and the latest depth frame. Unlike interpolation, our method does not use future frames.

\noindent
\textbf{Event-based 6D Pose Estimation.} 
Event cameras provide high-temporal-resolution and robustness to illumination~\cite{jeong2024towards, cho2024benchmark}, enabling event-only ego-motion~\cite{kim2016real,rebecq2017evo,rebecq2018ultimate, Awasthi2025MteventAM}, stereo and visual–inertial fusion~\cite{gehrig2021esvo,tesvo2023,esvio2023,zhou2021spatiotemporal}, and globally optimized SLAM~\cite{velometer2024,deio2024,greslam2025}. These strengths have motivated 6DoF pose estimation via geometric line tracking~\cite{lopez2024linepose,edopt2024icra,eventflow2024mm}, hybrid event–RGB pipelines~\cite{eventrgb2025}, marker-based LEDs~\cite{alm2024wacv,ebfiducial2021}, and stereo for spacecraft~\cite{stereo2025spacecraft}. Despite this progress, event-based 6D tracking remains constrained by small, object-specific datasets collected in controlled settings~\cite{epose2025,rojtberg2025ycbev,dubeau2020rgb} and a predominant focus on geometric formulations with few data-driven methods.

%% file: sec_cr/3_methods.tex
\section{Approach}

\textbf{Problem Formulation.}
Given a CAD model of a rigid object and known camera intrinsics, our goal is to estimate the current object pose \( \mathbf{T}_t = [\mathbf{R}_t\!\mid\!\mathbf{t}_t] \) in camera frame at an arbitrary time step \( t \), where \( t \) is a normalized time in \([0,1)\): \( \tau = 0 \) corresponds to the timestamp of the most recent depth frame, and \( \tau = 1 \) denotes the next depth frame in the future. $\mathbf{R}_t \in SO(3)$ and $\mathbf{t}_t \in \mathbb{R}^3$ represent its rotation and translation.

We assume access to an initial pose estimate 
$\mathbf{T}_{0}$, asynchronous event data, $E_{0,t} = \{ e_i = (\mathbf{x}_i, p_i, \tau_i) \mid 0 \le \tau_i < t \}$,  and depth measurement $D_0$. Each event $e_i$ is defined by its pixel location $\mathbf{x}_i = (x_i, y_i)^\top$, timestamp $\tau_i$, and polarity $p_i \in \{-1,1\}$, indicating the sign of the brightness change.

\noindent
\textbf{Method Overview.}
To achieve robust tracking in dynamic scenarios, we increase the temporal frequency of pose updates. 
While prior methods~\cite{megapose, wen2023foundationpose,  wen2023bundlesdf, wen2021bundletrack} typically infer poses only at timestamps where a depth frame is available and thereby constrained by the limited frame rate from a depth camera, EventTrack6D performs tracking updates at finer temporal intervals. As shown in Fig.~\ref{fig:overall_framework}, our framework enables inference at an arbitrary time \(t\) using two modules. The dual-modal reconstruction module predicts the current depth from the most recent depth frame and intervening events, while simultaneously reconstructing a dense intensity image. The pose-refinement module then uses the CAD model and the previous pose to estimate the pose at time \(t\) by matching against the reconstructed image and depth following a render-and-compare paradigm~\cite{wen2023foundationpose, wen2020se, nguyen2025gotrack, li2018deepim, labbe2020cosypose}.

\subsection{Dual-modal Reconstruction}

Given the most recent depth frame \(D_0\), the dual-modal reconstruction module processes two separate event streams with distinct roles. 
The long-range stream \(E_{0,t}\), accumulated since \(D_0\), provides motion cues primarily for depth reconstruction. 
The short-range stream \(E_{t-\Delta t,t}\), collected over the most recent interval, focuses on fine temporal details for intensity image reconstruction.
Using these inputs, the module reconstructs the  intensity image \(I_t\) and depth map \(D_t\) for the current timestamp when these images are missing due to their limited frame rate. The process is illustrated in Fig.~\ref{fig:overall_framework} (top-right).

\noindent
\textbf{Intensity 
Reconstruction.}
From the short-range event stream \(E_{t-\Delta t,t}\), we extract spatio–temporal features with an event encoder for image reconstruction \(\phi_E^{I}\):
\[
F^{E}_{t-\Delta t,t} = \phi_E^{I}\!\left(E_{t-\Delta t,t}\right).
\]

Then, we employ a ConvLSTM to aggregate temporal context and produce temporally integrated event features:
\[
\tilde{F}^{E}_{t-\Delta t, t},\, h_t = \mathrm{ConvLSTM}\!\left(F^{E}_{t-\Delta t,t},\, h_{t-\Delta t}\right),
\]
where \(h_{t-\Delta t}\) is the hidden state propagated from the previous time step.
Finally, the image decoder \(\psi_E^{I}\) fuses the aggregated event features to reconstruct the current intensity image:
\[
I_t = \psi_E^{I}\!(\tilde{F}^{E}_{t-\Delta t, t}).
\]

The recurrent hidden state \(h_t\) is propagated to the next time step, serving as temporal memory for the subsequent dual-modal reconstruction.

\noindent
\textbf{Depth Reconstruction.}
Given the most recent depth frame \(D_0\) and the long-range event stream \(E_{0,t}\),
we first extract motion-related features using the event and depth encoders \(\phi_E^{M}\) and \(\phi_D^{M}\):
\(F^{E,M}_{0,t} = \phi_E^{M}(E_{0,t})\) from the long-range events and
\(F^{D,M}_{0} = \phi_D^{M}(D_0)\) from the depth frame.

Then, we concatenate these motion-related features and apply a convolutional layer to generate the fused motion feature \(\tilde{F}^{M}_{0,t}\). Next, the initial motion predictor, $\psi^M$, generates a coarse motion field from the fused features as $
M^{D}_{0,t} = \psi^M\! (\tilde{F}^{M}_{0,t})$.
Using the initial motion field, we warp the previous depth features to the current time:
\begin{equation}
F^{D, M}_{0 \rightarrow t} = \mathrm{Warp}(F^{D, M}_{0},\, M^{D}_{0,t}).
\end{equation}

We then compute a cost volume via a correlation layer~\cite{sun2018pwc, jonschkowski2020matters, wen2025foundationstereo} between the warped depth features and the fused motion feature:
\begin{equation}
\mathcal{C}_{0,t} = \mathrm{Corr}\!\left(F^{D,M}_{0 \rightarrow t},\, \tilde{F}^{M}_{0,t}\right).
\end{equation}

Finally, we estimate a residual motion vector by concatenating the cost volume, the initial motion field, and the fused motion feature, followed by a convolutional layer, to obtain the refined motion field, $\tilde{M}^{D}_{0,t}$, by residual addition:
\begin{equation}
\begin{aligned}
\Delta M_{0,t}^D = \mathrm{Conv} & (\mathrm{Concat}(\mathcal{C}_{0,t},\, M^{D}_{0,t},\, \tilde{F}^{M}_{0,t})), \\
\tilde{M}^{D}_{0,t} &= M^{D}_{0,t} + \Delta M_{0,t}^D.
\end{aligned}
\end{equation}

While the refined motion field enables temporal propagation of past depth information, geometric correction remains necessary to handle changes in 3D structure over time.
To this end, we extract geometry-related features from long-range events 
and the depth frame using the encoders \(\phi_E^{G}\) and \(\phi_D^{G}\), producing 
\(F^{E,G}_{0,t} = \phi_E^{G}(E_{0,t})\) and \(F^{D,G}_{0} = \phi_D^{G}(D_0)\).
Using the refined motion field \(\tilde{M}^{D}_{0,t}\), we warp the geometry-related depth features to the current time:
\begin{equation}
F^{D,G}_{0 \rightarrow t} = \mathrm{Warp}(F^{D,G}_{0},\, \tilde{M}^{D}_{0,t}).
\end{equation}

Event cues \(F^{E,G}_{0,t}\) capture long-range motion but are inherently sparse and edge-dominated, which causes ambiguity for depth reconstruction in textureless regions or under large motion. 
To supply dense photometric context that encodes correspondences and regularizes geometry, we also use the temporally integrated event features from the image reconstruction stage \(\tilde{F}^{E}_{t-\Delta t,t}\). 
We then combine \(F^{E,G}_{0,t}\) with \(\tilde{F}^{E}_{t-\Delta t,t}\) to reconcile changes in 3D structure over time.
These representations are concatenated and refined, and the geometry module \(\psi^G\) produces the depth at time \(t\):
\begin{equation}
D_t = \psi^{G} (\mathrm{Concat}(\tilde{F}^{D,G}_{0 \rightarrow t},\, F^{D,G}_{0},\, F^{E,G}_{0,t},\, \tilde{F}^{E}_{t-\Delta t,t})).
\end{equation}

\subsection{6D Pose Refinement}

Dual-modal reconstruction predicts both the intensity image \(I_t\) and the depth map \(D_t\) at arbitrary timestamps. This allows pose tracking under large motion to be decomposed into a sequence of simpler subproblems with smaller motion.
\begin{equation}
\mathbf{T}_t
= \left( \prod_{k=1}^{N} \mathbf{T}_{\,k-1 \rightarrow k} \right)\mathbf{T}_0,
\qquad \text{where } N=\frac{t}{\Delta t}.
\end{equation}

Inspired by recent work in pose refinements~\cite{megapose, wen2023foundationpose}, the reconstructed intensity and depth images can be seamlessly integrated into a pose-refinement module that adopts a CAD-based render-and-compare paradigm to generalize to novel objects without retraining, as long as the object CAD model is given at test time.

We restrict pose updates to a region of interest (ROI) around the object. The crop is adapted from the previous pose estimate: its center is obtained by projecting the object origin onto the image plane, and its size is chosen to cover the object and its local context. Dual reconstruction is performed only within this ROI, reducing the computational cost of modality alignment.

The refinement module iteratively predicts a pose update that aligns rendered object views with the observed input. At each iteration, the current estimate is initialized from the previous pose,
\(\mathbf{T}_t = [R_t \mid t_t] \leftarrow [R_{t-\Delta t} \mid t_{t-\Delta t}]\),
and independently updated as
\begin{equation}
\begin{aligned}
R_t^{+} &= \Delta R\, R_t, \\
t_t^{+} &= t_t + \Delta t,
\end{aligned}
~
\begin{aligned}
\text{where } (\Delta R, \Delta t) &= \mathcal{R}\big(\mathbf{T}_t, I_t, D_t\big).
\end{aligned}
\end{equation}
$I_t$ and $D_t$ denote the predicted intensity image and depth map at time $t$, and $\mathcal{R}(\cdot)$ is the refinement operator.

\subsection{Objective Function}
\textbf{Dual-modal Reconstruction Loss.}
The reconstructed intensity image \(I_t\) and depth map \(D_t\) are supervised using ground-truth data.
For the image reconstruction, we apply a perceptual loss, LPIPS~\cite{zhang2018unreasonable}, that encourages photometric consistency with the ground-truth image \(I_t^{gt}\):
\begin{equation}
\mathcal{L}_{\text{img}} = \mathrm{LPIPS}(I_t,\, I_t^{gt}).
\end{equation}
For depth reconstruction, we applied \(L_1\) loss as: 
\begin{equation}
\mathcal{L}_{\text{depth}} = \| D_t - D_t^{gt} \|_1.
\end{equation}
The overall reconstruction objective is a weighted combination of both terms:
\begin{equation}
\mathcal{L}_{\text{recon}} = \lambda_I \mathcal{L}_{\text{img}} + \lambda_D \mathcal{L}_{\text{depth}},
\end{equation}
where \(\lambda_I\) and \(\lambda_D\) balance the contributions of photometric and geometric supervision. 

\noindent
\textbf{6D Pose Refinement Loss.} Pose refinement is optimized using an $L2$ loss: 
\begin{equation}
    \mathcal{L}_{pose}=\lambda_r\|\Delta R-\Delta R^\star\|_2  + \lambda_t\|\Delta t - \Delta t^\star \|_2
\end{equation}
where $\Delta R^t$, $\Delta t^*$ are ground truth and $\lambda_r$, $\lambda_t$ are weights.

%% file: sec_cr/4_dataset.tex
\section{Dataset Generation and Acquisition}

\begin{figure}[t]
    \centering
    \includegraphics[width=1.0\linewidth]{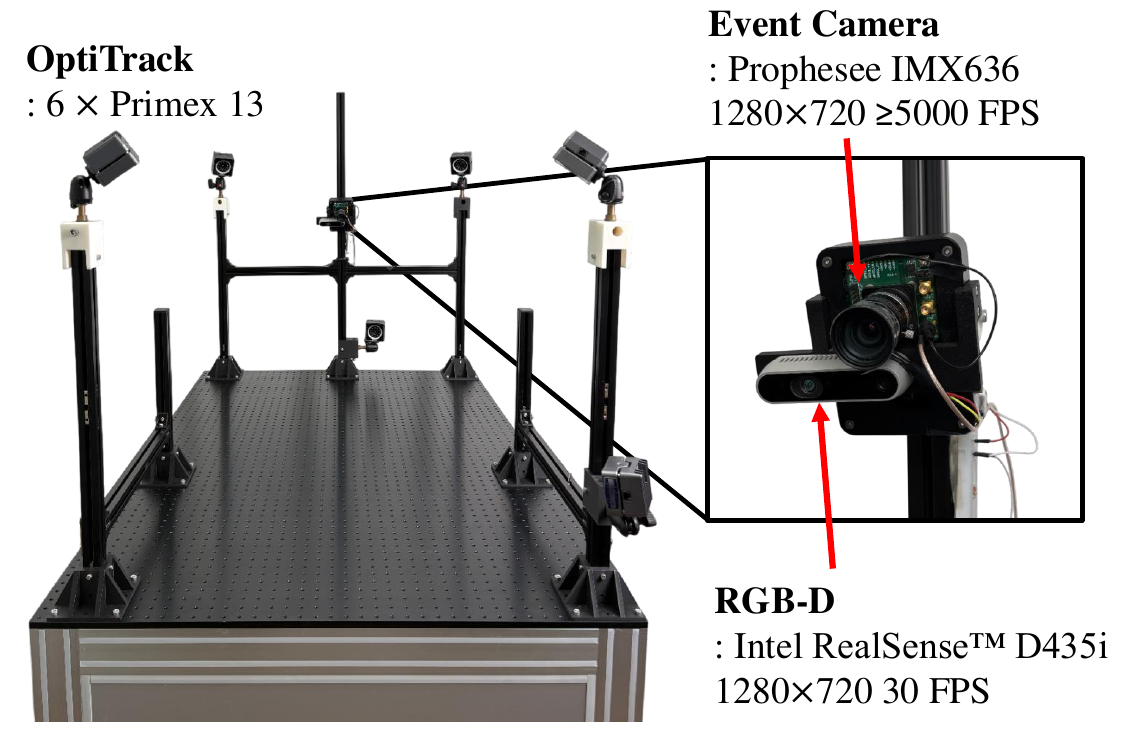}
    \vspace{-13pt}
    \caption{System designed for acquiring the Event6D dataset. The event camera, RGB-D camera, and motion capture system are all hardware-triggered, temporally synchronized and calibrated.}
    \label{fig:system}
    \vspace{-12pt}
\end{figure}

\subsection{EventBlender6D Dataset}
We present \textbf{EventBlender6D}, the first large-scale synthetic dataset 
for event-based object pose estimation and tracking. 
We build our pipeline with BlenderProc~\cite{denninger2020blenderproc}, enabling high-frame-rate RGB rendering and accurate annotations using Google Scanned Objects~\cite{downs2022google}.
An event simulator~\cite{rebecq2018esim} is then applied to the rendered sequences to produce 
synthetic event streams.
The objects synthesized in the EventBlender6D dataset are disjoint from those in the evaluation datasets, Event6D and EventHO3D.

\subsection{Event6D Dataset}

We propose \textbf{Event6D} dataset, a real-world event-based 6D object pose dataset. 
As shown in Fig.~\ref{fig:system}, motion capture system is employed to obtain accurate annotations. 
The RGB-D camera, event camera, and motion capture system are hardware-triggered, 
ensuring precise temporal synchronization. 
Event6D provides high-quality 6D pose ground truth at 120 FPS, enabling reliable benchmarking of event-based high-frame-rate 6D object pose tracking methods. The RGB and depth streams were recorded at 30 FPS, following existing datasets~\cite{hot3d, chao2021dexycb, rojtberg2025ycbev}. Our Event6D dataset includes a subset of objects from YCB~\cite{xiang2017posecnn, calli2015ycb} and  HOGraspNet~\cite{cho2024dense} datasets with existing CAD models. To incorporate novel objects, we captured real objects using a 3D scanner and generated corresponding CAD meshes. The dataset includes diverse and challenging scenarios. Further details are provided in the supplementary material.

\subsection{EventHO3D Dataset}
The HO3D dataset~\cite{hampali2020honnotate} provides video sequences commonly used for pose tracking evaluation. To assess the generalization of our method, we generate an event-based counterpart, EventHO3D, by simulating events using ESIM~\cite{rebecq2018esim}.

%% file: sec_cr/5_experiments.tex
\definecolor{Ours}{RGB}{255,235,225}

\section{Experiments}

\begin{table*}[t]
\caption{Comparison of event-based 6D object pose tracking methods on Event6D dataset, evaluated against ground-truth poses at 120 FPS. The event camera operates asynchronously at high frame rates, while RGB or depth is only available at 30 FPS. Therefore, conventional RGB or RGB-D-based methods cannot be evaluated under the 120 FPS ground-truths. 
The runtime is measured on preprocessed data, with a patch size of 160 × 160 corresponding to the region of interest in FP.
$\dagger$ denotes that the model is trained for event inputs and bracket (·) in FPS indicates the runtime when applied TensorRT.} 
\vspace{-4pt}
\centering
\renewcommand{\arraystretch}{1.05}
\resizebox{.99\textwidth}{!}{
\setlength{\tabcolsep}{6pt}
\begin{tabular}{c||ccc|ccc|ccc|ccc|ccc}
\thickhline
\multirow{1}{*}{Approach}
 & \multicolumn{3}{c|}{E2VID~\cite{rebecq2019high} + MG~\cite{megapose}} 
 & \multicolumn{3}{c|}{E2VID~\cite{rebecq2019high} + FP~\cite{wen2023foundationpose}} 
 & \multicolumn{3}{c|}{ETAP~\cite{hamann2025etap}}
& \multicolumn{3}{c|}{\multirow{1}{*}{FP$^{\dagger}$~\cite{wen2023foundationpose}}}
  & \multicolumn{3}{c}{\cellcolor{grey}\textbf{EventTrack6D (Ours)}} 
  \\
\hline
Modality & \multicolumn{3}{c|}{Event + Depth} & \multicolumn{3}{c|}{Event + Depth} & \multicolumn{3}{c|}{Event + Depth} & \multicolumn{3}{c|}{Event + Depth} & \multicolumn{3}{c}{\cellcolor{grey} Event + Depth} \\
Metric & ADD-S & ADD & AR &ADD-S & ADD & AR &ADD-S & ADD & AR & ADD-S & ADD & AR & \cellcolor{grey} ADD-S & \cellcolor{grey} ADD & \cellcolor{grey} AR \\
\hline
banana & 0.00 & 0.00 & 0.17 & 24.92 & 9.65 & 25.38 & 6.09 & 1.54 & 7.11 & 13.99 & 1.70 & 9.43 & \cellcolor{grey} 43.46 & \cellcolor{grey} 16.26 & \cellcolor{grey} 41.83 \\
bowl & 4.43 & 0.00 & 17.41 & 46.20 & 0.57 & 77.11 & 48.20 & 1.25 & 75.21 & 2.70 & 0.38 & 8.98 & \cellcolor{grey} 54.86 & \cellcolor{grey} 0.34 & \cellcolor{grey} 85.65 \\
cracker & 0.00 & 0.00 & 0.03 & 36.65 & 28.42 & 42.05 & 1.17 & 1.04 & 1.31 & 45.53 & 27.28 & 60.42 & \cellcolor{grey} 74.65 & \cellcolor{grey} 62.44 & \cellcolor{grey} 89.19 \\
drill & 0.16 & 0.06 & 1.56 & 4.24 & 2.67 & 5.10 & 1.02 & 0.53 & 1.20 & 39.04 & 6.83 & 21.66 & \cellcolor{grey} 66.13 & \cellcolor{grey} 38.58 & \cellcolor{grey} 64.94 \\
hammer & 35.17 & 16.16 & 35.55 & 57.66 & 40.34 & 62.46 & 39.61 & 27.64 & 42.82 & 23.32 & 3.20 & 12.47 & \cellcolor{grey} 53.84 & \cellcolor{grey} 39.68 & \cellcolor{grey} 57.66 \\
marker & 12.04 & 0.83 & 17.32 & 0.17 & 0.17 & 0.17 & 0.17 & 0.17 & 0.17 & 12.33 & 1.02 & 8.44 & \cellcolor{grey} 62.36 & \cellcolor{grey} 28.01 & \cellcolor{grey} 65.92 \\
mouse & 2.10 & 0.43 & 10.74 & 25.48 & 3.31 & 63.32 & 0.13 & 0.13 & 0.36 & 8.60 & 0.81 & 9.89 & \cellcolor{grey} 34.54 & \cellcolor{grey} 4.93 & \cellcolor{grey} 70.48 \\
mug & 0.00 & 0.00 & 0.24 & 21.43 & 6.34 & 27.79 & 1.96 & 0.80 & 4.16 & 3.46 & 0.29 & 2.58 & \cellcolor{grey} 38.74 & \cellcolor{grey} 8.08 & \cellcolor{grey} 39.30 \\
mustard & 0.59 & 0.31 & 1.10 & 70.79 & 49.66 & 81.07 & 5.20 & 3.62 & 5.42 & 26.38 & 4.42 & 21.56 & \cellcolor{grey} 82.21 & \cellcolor{grey} 63.26 & \cellcolor{grey} 89.79 \\
pitcher & 12.83 & 4.52 & 17.70 & 72.35 & 43.29 & 74.51 & 8.44 & 5.12 & 8.19 & 58.15 & 17.66 & 47.99 & \cellcolor{grey} 71.47 & \cellcolor{grey} 42.86 & \cellcolor{grey} 81.02 \\
scrub & 3.93 & 1.26 & 3.98 & 76.67 & 54.53 & 82.57 & 1.21 & 1.01 & 1.44 & 61.01 & 22.19 & 46.54 & \cellcolor{grey} 84.55 & \cellcolor{grey} 65.97 & \cellcolor{grey} 91.63 \\
spam & 1.28 & 0.32 & 18.60 & 49.74 & 23.66 & 71.36 & 8.47 & 4.01 & 12.59 & 27.44 & 8.03 & 48.64 & \cellcolor{grey} 49.03 & \cellcolor{grey} 24.40 & \cellcolor{grey} 77.01 \\
spatula & 2.07 & 0.99 & 4.23 & 47.97 & 33.39 & 50.07 & 2.68 & 2.02 & 2.98 & 5.07 & 2.23 & 4.22 & \cellcolor{grey} 0.10 & \cellcolor{grey} 0.10 & \cellcolor{grey} 0.10 \\
wine & 0.18 & 0.18 & 0.27 & 0.18 & 0.18 & 1.23 & 0.18 & 0.18 & 0.19 & 16.66 & 0.94 & 19.87 & \cellcolor{grey} 17.87 & \cellcolor{grey} 0.48 & \cellcolor{grey} 26.21 \\

\hline
MEAN & 6.78 & 2.12 & 9.02 & 37.24 & 16.97 & 48.72 & 7.77 & 2.22 & 11.42 & 22.93 & 4.31 & 25.72 & \cellcolor{grey}\textbf{52.79} & \cellcolor{grey}\textbf{25.26} & \cellcolor{grey}\textbf{64.38} \\
FPS (Hz) & \multicolumn{3}{c|}{10.92} & \multicolumn{3}{c|}{79.37} & \multicolumn{3}{c|}{1.56} & \multicolumn{3}{c|}{\textbf{108.70}} & \multicolumn{3}{c}{\cellcolor{grey} 50.19 (128.04)}\\
\thickhline
\end{tabular}
}
\vspace{-10pt}
\label{tab:high_fps_comparison}
\end{table*}

\noindent \textbf{Experimental Setup.}
We train our model on the synthetic EventBlender6D dataset and evaluate it (without any additional training or fine-tuning) on the real Event6D and synthetic EventHO3D datasets under the novel object setting~\cite{hodan2024bop, wen2023foundationpose, moon2025co}, where all evaluation objects are unseen during training.
Following \cite{wen2023foundationpose}, we assume that only the first frame of 6D pose is provided and evaluate long-term tracking robustness without re-initialization.

Evaluation is conducted for two different settings: 120 FPS and 30 FPS. 
In the 120 FPS setting, where RGB-D data (captured at 30 FPS) are unavailable, we compare our method against event-based baselines under high-frame-rate inference.
For fair comparison with conventional RGB-D methods, we evaluate our approach at 30 FPS.

\begin{table*}[t]
\caption{Comparison of event-based 6D pose tracking methods on Event6D dataset at 30 FPS. All methods are evaluated at RGB/depth frame intervals for fair comparison with RGB-based methods. $\dagger$ denotes models trained on event inputs.} \vspace{-5pt}
\centering
\renewcommand{\arraystretch}{1.2}
\resizebox{.99\textwidth}{!}{
\setlength{\tabcolsep}{4pt}
\begin{tabular}{c||ccc|ccc|ccc|ccc|ccc|ccc}
\thickhline
\multirow{1}{*}{Approach}
 & \multicolumn{3}{c|}{\multirow{1}{*}{MG~\cite{megapose}}}  &  \multicolumn{3}{c|}{\multirow{1}{*}{FP~\cite{wen2023foundationpose}}}  & \multicolumn{3}{c|}{E2VID~\cite{rebecq2019high} + MG~\cite{megapose}} & \multicolumn{3}{c|}{E2VID~\cite{rebecq2019high} + FP~\cite{wen2023foundationpose}} & \multicolumn{3}{c|}{\multirow{1}{*}{FP$^{\dagger}$~\cite{wen2023foundationpose}}}  & \multicolumn{3}{c}{\cellcolor{grey}\textbf{EventTrack6D (Ours)}} \\
\hline
Modality & \multicolumn{3}{c|}{RGB + Depth} & \multicolumn{3}{c|}{RGB + Depth} & \multicolumn{3}{c|}{Event + Depth} & \multicolumn{3}{c|}{Event + Depth} & \multicolumn{3}{c|}{Event + Depth} & \multicolumn{3}{c}{\cellcolor{grey} Event + Depth} \\
Metric & ADD-S & ADD & AR & ADD-S & ADD & AR &ADD-S & ADD & AR &ADD-S & ADD & AR & ADD-S & ADD & AR & \cellcolor{grey} ADD-S & \cellcolor{grey} ADD & \cellcolor{grey} AR    \\
\hline
banana  & 24.87 & 11.78 & 22.56 & 20.52 & 10.23 & 17.61 & 0.00 & 0.00 & 0.26 & 16.32 & 5.64 & 14.91 & 24.96 & 2.11 & 15.39 & \cellcolor{grey} 35.68 & \cellcolor{grey} 13.58 & \cellcolor{grey} 33.21 \\
bowl  & 1.13 & 0.47 & 2.08 & 36.14 & 0.97 & 21.18 & 0.78 & 0.00 & 11.04 & 48.43 & 1.07 & 73.95 & 3.80 & 0.38 & 10.31 & \cellcolor{grey} 63.19 & \cellcolor{grey} 1.41 & \cellcolor{grey} 88.18 \\
cracker  & 5.45 & 2.66 & 5.68 & 32.05 & 24.65 & 40.10 & 0.00 & 0.00 & 0.15 & 69.29 & 52.49 & 79.59 & 53.20 & 34.33 & 64.55 & \cellcolor{grey} 76.40 & \cellcolor{grey} 65.88 & \cellcolor{grey} 90.06 \\
drill  & 27.75 & 9.78 & 25.88 & 14.33 & 9.89 & 13.58 & 0.53 & 0.23 & 0.90 & 8.00 & 4.42 & 7.60 & 39.91 & 4.80 & 20.97 & \cellcolor{grey} 64.24 & \cellcolor{grey} 42.68 & \cellcolor{grey} 64.90 \\
hammer & 11.56 & 5.38 & 13.24 & 3.19 & 1.99 & 2.60 & 18.10 & 2.66 & 27.13 & 43.11 & 28.81 & 46.17 & 27.85 & 3.56 & 13.52 & \cellcolor{grey} 57.96 & \cellcolor{grey} 44.20 & \cellcolor{grey} 59.81 \\
marker  & 7.55 & 1.24 & 5.23 & 6.31 & 2.86 & 6.84 & 2.57 & 0.68 & 6.51 & 51.38 & 19.04 & 50.19 & 14.05 & 1.37 & 8.41 & \cellcolor{grey} 49.94 & \cellcolor{grey} 23.68 & \cellcolor{grey} 49.28 \\
mouse  & 0.00 & 0.00 & 0.72 & 1.04 & 0.52 & 0.92 & 0.81 & 0.00 & 6.13 & 26.25 & 3.85 & 57.37 & 10.82 & 1.23 & 10.58 & \cellcolor{grey} 34.30 & \cellcolor{grey} 4.06 & \cellcolor{grey} 70.95 \\
mug  & 3.57 & 1.58 & 3.82 & 1.11 & 0.58 & 1.12 & 0.00 & 0.00 & 0.19 & 4.28 & 1.70 & 6.53 & 4.57 & 0.29 & 2.94 & \cellcolor{grey} 43.35 & \cellcolor{grey} 9.58 & \cellcolor{grey} 39.08 \\
mustard  & 0.94 & 0.51 & 2.00 & 5.74 & 4.38 & 6.72 & 0.00 & 0.00 & 0.22 & 63.62 & 38.70 & 72.22 & 27.89 & 5.71 & 20.56 & \cellcolor{grey} 85.10 & \cellcolor{grey} 68.15 & \cellcolor{grey} 91.81 \\
pitcher & 20.39 & 12.38 & 20.69 & 55.66 & 40.23 & 59.25 & 0.93 & 0.00 & 8.28 & 69.00 & 37.25 & 69.72 & 63.28 & 23.83 & 51.28 & \cellcolor{grey} 77.81 & \cellcolor{grey} 54.42 & \cellcolor{grey} 85.17 \\
scrub & 4.60 & 2.30 & 5.32 & 5.76 & 4.40 & 5.22 & 2.25 & 0.49 & 2.24 & 76.20 & 53.88 & 82.02 & 67.73 & 31.41 & 53.54 & \cellcolor{grey} 89.65 & \cellcolor{grey} 77.70 & \cellcolor{grey} 94.10 \\
spam &  2.86 & 1.74 & 5.48 & 14.85 & 11.04 & 17.50 & 0.00 & 0.00 & 9.08 & 46.05 & 22.21 & 68.32 & 37.48 & 13.23 & 55.20 & \cellcolor{grey} 60.11 & \cellcolor{grey} 36.37 & \cellcolor{grey} 81.15 \\
spatula & 1.10 & 0.45 & 0.66 & 14.74 & 10.13 & 13.37 & 1.15 & 0.39 & 2.99 & 0.10 & 0.10 & 0.18 & 6.24 & 3.50 & 4.48 & \cellcolor{grey} 0.38 & \cellcolor{grey} 0.38 & \cellcolor{grey} 0.38 \\
wine &  1.07 & 0.00 & 0.80 & 0.73 & 0.73 & 0.73 & 0.73 & 0.73 & 0.47 & 0.36 & 0.18 & 2.43 & 14.68 & 1.51 & 19.12 & \cellcolor{grey} 1.65 & \cellcolor{grey} 0.73 & \cellcolor{grey} 1.80 \\
\hline
MEAN & 11.08 & 5.59 & 10.49 & 16.63 & 7.98 & 19.25 & 2.05 & 0.19 & 5.24 & 35.45 & 16.03 & 44.92 & 26.21 & 5.87 & 27.53 & \cellcolor{grey}\textbf{56.08} & \cellcolor{grey}\textbf{29.40} & \cellcolor{grey}\textbf{63.71} \\
\thickhline
\end{tabular}
}
\vspace{-10pt}
\label{tab:low_fps_comparison}
\end{table*}

\begin{figure*}[h!]
    \centering
    \includegraphics[width=.96\textwidth]{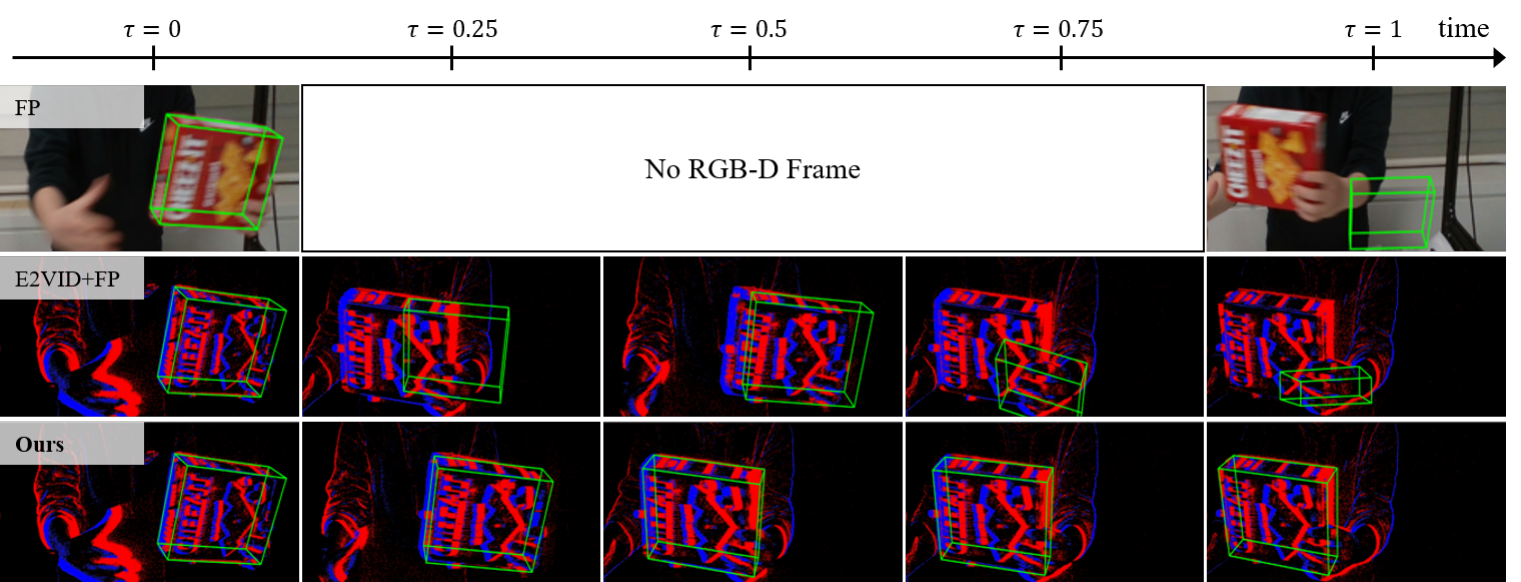}
    \vspace{-5pt}
    \caption{Qualitative comparison of 6D object tracking at 120 FPS on the Event6D dataset. Original FoundationPose(FP)~\cite{wen2023foundationpose} assumes  RGB-D input and thus cannot be applied to a high frame rate setting. Note that for $\tau = 0.25, 0.5, 0.75$, ours utilizes its reconstructed depth rather than sensor-captured depth.
    }
    \label{fig:main_qual}
    \vspace{-10pt}
\end{figure*}

\noindent
\textbf{Evaluation Metrics.}
We report standard object pose metrics including the area under the curve (AUC) of ADD and ADD-S~\cite{hinterstoisser2012model, xiang2017posecnn}, as well as the Average Recall (AR)~\cite{nguyen2025bop} of Visible Surface Discrepancy (VSD), Maximum Symmetry-Aware Surface Distance (MSSD), and Maximum Symmetry-Aware Projection Distance (MSPD).
Following previous studies~\cite{wen2023foundationpose, nguyen2025bop}, we use a threshold of 0.1 times the object diameter for ADD and ADD-S, and varying thresholds for AR metrics.
We measure inference time on an NVIDIA RTX A6000 GPU, enforcing CPU–GPU synchronization following prior work~\cite{Gehrig_2023_CVPR}.

\noindent

\textbf{Event-based Baselines.} 
To the best of our knowledge, there are no existing learning–based event-driven methods for novel 6D object pose tracking.
We therefore compare two categories of baselines: (1) frame-based methods using event-to-image conversion and (2) event-based methods that operate directly on events.
For the first category, we adapt state-of-the-art (SOTA) RGB-D-based 6D pose tracking methods, MegaPose (MG)~\cite{megapose} and FoundationPose (FP)~\cite{wen2023foundationpose}, using E2VID~\cite{e2vid} to convert event streams into images compatible with their pipelines.
For the second category, we use event-based baselines that operate directly on event and depth. ETAP~\cite{hamann2025etap} tracks query points sampled from the CAD surface and estimates the 6D pose by fitting a rigid transformation~\cite{besl1992method_ICP,lepetit2009ep}. In addition, we fine-tune FP on the EventBlender6D dataset to support event inputs.
All event-based baselines support both event-only and event-depth input. Details are in the supplementary material.

\noindent
\textbf{RGB-based Baselines.} 
We also compare against the RGB-D SOTA pose-tracking methods MegaPose and FoundationPose, under 30 FPS settings only.

\subsection{Comparison on Event6D dataset}

\noindent
\textbf{Evaluation under the 120 FPS setting.}
In Table~\ref{tab:high_fps_comparison}, we evaluate baselines for event-based methods at 120 FPS. Since depth is captured at 30 FPS (every fourth frame), baselines use depth and events when available, and events only at intermediate frames.

E2VID~\cite{e2vid} + MG~\cite{megapose} and E2VID + FP~\cite{wen2023foundationpose} struggle with temporally sparse depth, leading to unstable pose tracking under rapid motion, as shown in \Fref{fig:main_qual}.
ETAP~\cite{hamann2025etap}-based point tracking with ICP~\cite{besl1992method_ICP} and PnP~\cite{lepetit2009ep} also degrades under dynamic motion and occlusion, which limits overall performance. For fine-tuned FP using event inputs, the modality mismatch between CAD renderings and event streams, together with intermittent depth measurements, results in suboptimal performance.

Our method reconstructs intensity and depth at arbitrary timestamps, producing CAD-aligned observations that enable 6D pose tracking and robust performance across objects. 
The model runs at 50 FPS without optimization and exceeds 120 FPS with TensorRT.

\noindent
\textbf{Evaluation under the 30 FPS setting.}
In this section, we compare our method against RGB-D pose tracking baselines FP~\cite{wen2023foundationpose}, and MG~\cite{megapose} at 30 FPS (\Tref{tab:low_fps_comparison}), where RGB-D data are available. The event-based methods perform pose tracking at 120 FPS with the same configuration as in Table~\ref{tab:high_fps_comparison}, but the pose estimates are evaluated at 30 FPS.
Consistent with 120 FPS results (\Tref{tab:high_fps_comparison}), EventTrack6D outperforms other event-depth baselines. 
 Moreover, our method surpasses strong RGB-D baselines, MG~\cite{megapose} and FP~\cite{wen2023foundationpose}.
 Both strong RGB-D baselines often fail when faced with large inter-frame motion or severe motion blur, leading to unstable pose tracking.

\begin{table}[t]
\caption{Comparison on the EventHO3D dataset. 
$\dagger$ denotes that the model is trained for event inputs.
}
\vspace{-5pt}
\centering
\renewcommand{\arraystretch}{1.2}
\resizebox{.478\textwidth}{!}{
\setlength{\tabcolsep}{2pt}
\begin{tabular}{c||ccc|ccc|ccc}
\thickhline
\multirow{1}{*}{Method}
 & \multicolumn{3}{c|}{E2VID~\cite{rebecq2019high} + FP~\cite{wen2023foundationpose}} & \multicolumn{3}{c|}{\multirow{1}{*}{FP$^{\dagger}$~\cite{wen2023foundationpose}}}  & \multicolumn{3}{c}{\cellcolor{grey}\textbf{EventTrack6D (Ours)}} \\
\hline
Modality &  \multicolumn{3}{c|}{Event + Depth} & \multicolumn{3}{c|}{Event + Depth} & \multicolumn{3}{c}{\cellcolor{grey} Event + Depth} \\
Metric & ADD-S & ADD & AR & ADD-S & ADD & AR & \cellcolor{grey} ADD-S & \cellcolor{grey} ADD & \cellcolor{grey} AR \\
\hline
AP10  & 52.60 & 34.86 & 46.75 & 10.41 & 0.63 & 4.76 & \cellcolor{grey} 88.66 & \cellcolor{grey} 77.06 & \cellcolor{grey} 90.08 \\
AP11  & 82.55 & 57.65 & 83.16 & 49.62 & 13.89 & 38.62 & \cellcolor{grey} 42.51 & \cellcolor{grey} 29.09 & \cellcolor{grey} 40.52 \\
AP12  & 70.74 & 26.00 & 50.48 & 64.96 & 24.05 & 55.36 & \cellcolor{grey} 85.29 & \cellcolor{grey} 64.90 & \cellcolor{grey} 82.54 \\
AP13  & 80.93 & 51.59 & 77.83 & 59.57 & 17.37 & 41.34 & \cellcolor{grey} 78.17 & \cellcolor{grey} 42.13 & \cellcolor{grey} 66.02 \\
AP14  & 53.52 & 42.28 & 51.67 & 12.06 & 6.44 & 10.75 & \cellcolor{grey} 85.09 & \cellcolor{grey} 67.78 & \cellcolor{grey} 83.06 \\
SM1   & 41.95 & 33.78 & 45.31 & 0.64 & 0.64 & 0.64 & \cellcolor{grey} 85.75 & \cellcolor{grey} 74.99 & \cellcolor{grey} 85.79 \\
SB11  & 45.08 & 36.44 & 47.24 & 43.21 & 5.93 & 23.50 & \cellcolor{grey} 88.17 & \cellcolor{grey} 75.54 & \cellcolor{grey} 90.82 \\
SB13  & 44.08 & 37.54 & 46.37 & 38.30 & 7.62 & 36.82 & \cellcolor{grey} 92.00 & \cellcolor{grey} 83.96 & \cellcolor{grey} 95.42 \\
MPM10 & 65.58 & 55.11 & 70.13 & 36.48 & 9.92 & 23.80 & \cellcolor{grey} 40.31 & \cellcolor{grey} 33.21 & \cellcolor{grey} 43.03 \\
MPM11 & 53.76 & 29.30 & 45.18 & 2.05 & 0.64 & 1.11   & \cellcolor{grey} 43.03 & \cellcolor{grey} 36.60 & \cellcolor{grey} 46.90 \\
MPM12 & 32.73 & 25.99 & 32.98 & 58.35 & 12.83 & 43.96 & \cellcolor{grey} 43.94 & \cellcolor{grey} 38.27 & \cellcolor{grey} 46.78 \\
MPM13 & 91.30 & 84.18 & 95.18 & 64.50 & 34.79 & 57.07 & \cellcolor{grey} 48.42 & \cellcolor{grey} 39.77 & \cellcolor{grey} 49.59 \\
MPM14 & 35.33 & 29.31 & 34.63 & 36.14 & 2.27 & 11.25 & \cellcolor{grey} 43.08 & \cellcolor{grey} 34.79 & \cellcolor{grey} 45.46 \\
\hline
MEAN 
 & 57.10 & 41.13 & 56.77 
 & 35.31 & 8.83 & 27.64 
 & \cellcolor{grey} \textbf{64.75} 
 & \cellcolor{grey} \textbf{50.95} 
 & \cellcolor{grey} \textbf{66.27} \\
\thickhline
\end{tabular}
}
\vspace{-3pt}
\label{tab:ho3d_comparison}
\end{table}

\noindent
\noindent
\textbf{Qualitative Results.}
Figure~\ref{fig:main_qual} shows 120 FPS pose-tracking results. The RGB-D baseline fails under large motions because it is unable to access depth frames for pose tracking due to limited frame rates. The hybrid E2VID~\cite{e2vid}+FP~\cite{wen2023foundationpose} approach enables more frequent updates via image reconstruction but still degrades when depth is unavailable. In contrast, EventTrack6D reconstructs observations at arbitrary timestamps, enabling robust tracking across diverse motion conditions.

\begin{figure*}[t]
    \centering
   \includegraphics[width=.94\textwidth]{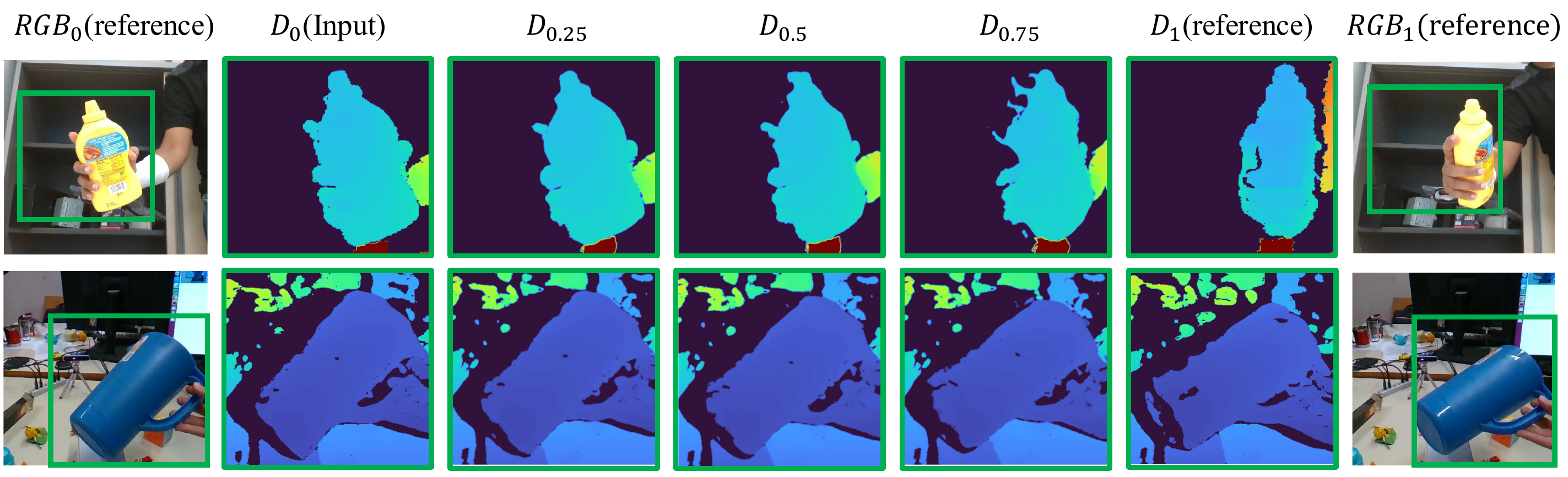}
    \vspace{-10pt}
    \caption{Qualitative depth-reconstruction results on depth-absent intervals.
The future depth \(\,D_1\,\) is provided solely for reference and is not used by the method.
Despite dynamic motion, our approach reconstructs depth images that preserve coherent object structure and align with the object motion, providing geometric guidance for downstream pose tracking. 
}
    \label{fig:depth_extra_qual}
    \vspace{-10pt}
\end{figure*}

\subsection{Comparison on EventHO3D Dataset}
We further evaluate the baselines on a different domain, the EventHO3D dataset, which differs from Event6D in both motion distributions and sensor settings, providing an additional test of domain generalization.

As shown in \Tref{tab:ho3d_comparison}, EventTrack6D generalizes across datasets, even when trained solely on the synthetic EventBlender6D dataset. For comparison, we also report results for other event-based baselines, evaluated under a 30-FPS protocol following Table~\ref{tab:low_fps_comparison}. 
Because EventHO3D exhibits more moderate motion than Event6D, tracking failures are less frequent even when depth is missing at intermediate timesteps. Nevertheless, by jointly reconstructing intensity and depth, EventTrack6D outperforms all baselines.

\begin{table}[t]
\caption{Ablation study of the dual-modal reconstruction.}
\vspace{-5pt}
\centering
\setlength{\tabcolsep}{6.7pt}
\resizebox{.478\textwidth}{!}{
\begin{tabular}{cc||ccc}
\thickhline
Depth Recon. & Image Recon. & \multirow{1}{*}{ADD-S$\uparrow$} & \multirow{1}{*}{ADD$\uparrow$} & \multirow{1}{*}{AR$\uparrow$} \\
\hline
& & 18.45&3.29&20.07
\\
 \checkmark & & 28.67	&4.75&	29.08
\\
& \checkmark & 30.53 & 13.79 & 44.99
\\
\checkmark & \checkmark & 52.79 & 25.26 & 64.38 \\
\thickhline
\end{tabular}
}
\vspace{-10pt}
\label{tab:ablation_dual}
\end{table}

\subsection{Ablation Study and Analysis}

\noindent
\textbf{Dual-modal Reconstruction.}
We conduct an ablation study on the 120 FPS Event6D dataset, as summarized in Table~\ref{tab:ablation_dual}. In the baseline configuration, both depth and image reconstruction are removed: depth is used only when available at 30 FPS, and the refinement module is trained on EventBlenderProc to handle event inputs directly without image reconstruction.
This setup suffers from limited geometric information between sparse depth frames and a modality mismatch between CAD renderings and event observations, ultimately leading to degraded performance.

When only depth reconstruction is added, tracking at arbitrary timestamps becomes feasible by injecting geometric cues, yet training remains challenging due to the persistent mismatch between event inputs and CAD renderings. When only image reconstruction is considered, photometric alignment improves, but the lack of continuous geometric information limits robustness under dynamic motion.

In contrast, our dual-modal reconstruction produces observations that align well with CAD renderings at arbitrary timestamps, providing both geometric and photometric cues, thereby achieving consistently superior performance across diverse motion conditions.

\noindent
\textbf{Depth Reconstruction.}
Table~\ref{tab:ablation_depth} summarizes ablations on the depth reconstruction module. Incorporating image features provides dense visual context that complements sparse event features, improving reconstruction by adding foreground and texture information. Motion vectors capture object-centric dynamics and enable accurate estimation of geometric changes under diverse motion patterns. Fusing image features with motion vectors improves depth predictions that more accurately represent real-world geometry.

Figure~\ref{fig:depth_extra_qual} shows reconstruction results at intervals between depth observations($\tau=0$ , $\tau=1$). The module produces depth maps that are consistent with available observations and pixel-aligned with RGB images. Such realistic depth is input to the pose refinement stage and significantly enhances event-based 6D pose tracking.

\begin{table}[t]
\caption{Ablation study of depth-reconstruction components.}
\vspace{-6pt}
\centering
\setlength{\tabcolsep}{6.7pt}
\resizebox{.478\textwidth}{!}{
\begin{tabular}{cc||ccc}
\thickhline
Motion Vector & Image Feature & \multirow{1}{*}{ADD-S$\uparrow$} & \multirow{1}{*}{ADD$\uparrow$} & \multirow{1}{*}{AR$\uparrow$} \\
\hline
& & 36.03	&17.52&43.60\\
& \checkmark & 42.26	&19.32	&48.99
\\
\checkmark & &41.88	&17.61&	50.54
\\
\checkmark & \checkmark & 52.79 & 25.26 & 64.38 \\
\thickhline
\end{tabular}
}
\vspace{-16pt}
\label{tab:ablation_depth}
\end{table}

%% file: sec_cr/6_conclusion.tex
\section{Conclusion}
In this paper, we explore the problem of event-based 6D object pose tracking.
Due to the lack of large-scale datasets for training and evaluation, we introduce three datasets: EventBlender6D, EventHO3D, and Event6D. Moreover, we propose the EventTrack6D framework for novel 6D object pose tracking.
Our efficient event-aware design processes 6D pose tracking at 128 FPS.
Our experiments demonstrate strong generalization capability in 6D object pose tracking tasks, effectively handling the unique characteristics of event cameras. We believe this work will foster further research on event-based perception and high-speed 6D pose tracking.